\documentclass[10pt,compsoc]{IEEEtran}
\usepackage{amsmath,amsfonts}
\usepackage{algorithmic}
\usepackage{algorithm}
\usepackage{array}

\usepackage{textcomp}
\usepackage{stfloats}
\usepackage{bbding}
\usepackage{url}
\usepackage{verbatim}
\usepackage{graphicx}
\usepackage{float}
\usepackage{balance} 

\usepackage{hyperref}
\usepackage{graphicx}

\usepackage{pifont}
\usepackage{cite}
\usepackage{color}
\usepackage{xcolor}
\usepackage{diagbox}
\usepackage{enumitem}
\usepackage{colortbl}

\hypersetup{colorlinks=true,
	linkcolor=magenta,
	anchorcolor=blue,
	citecolor=blue,
	urlcolor=magenta
}
\usepackage{makecell}
\usepackage{booktabs}
\usepackage{multirow}
\usepackage{orcidlink}
\usepackage[table]{xcolor}
\usepackage{tikz}
\usepackage{array}
\usepackage{ulem}
\usepackage{ragged2e}
\usepackage{utfsym}
\usepackage{wasysym}
\usepackage{amssymb}
\usepackage{threeparttable}
\usepackage{capt-of} 
\usepackage{multirow}
\definecolor{t1}{RGB}{200,0,0}
\definecolor{t2}{RGB}{0,0,200}
\newcommand{\tone}[1]{\textcolor{t1}{\textbf{#1}}}
\newcommand{\ttwo}[1]{\textcolor{t2}{\textbf{#1}}}
\usepackage{makecell}

\usepackage[caption=false,font=footnotesize,labelfont=sf,textfont=sf]{subfig}
\newcommand{\etc}{\textit{etc}. }
\newcommand{\ie}{\textit{i}.\textit{e}., }

\newtheorem{proposition}{\textbf{Proposition}}

\begin{document}
	\bstctlcite{IEEEexample:BSTcontrol}
    
    \title{MMATR-X: High-Resolution Optical-SAR Joint Pretraining via Conditioned Alignment and Degraded-Target Prediction}

    \author{Bowen Peng, Yongxiang Liu, Jie Zhou, Xiaodong Chen, Tianpeng Liu, Xiaogang Yu, Li Liu
			
        \IEEEcompsocitemizethanks{
            \IEEEcompsocthanksitem This work was supported by the New Generation Artificial Intelligence-National Science and Technology Major Project under Grant 2026ZD0126400, the National Natural Science Foundation of China (NSFC) under Grant 62376283 and 62531026, the Innovation Research Foundation of National University of Defense Technology (JS2023-03), and the Fundamental and Interdisciplinary Disciplines Breakthrough Plan of the Ministry of Education of China (JYB2025XDXM110). \textit{(Corresponding authors: Yongxiang Liu and Li Liu.)}
            \IEEEcompsocthanksitem Bowen Peng, Yongxiang Liu, Jie Zhou, Xiaodong Chen, Tianpeng Liu, and Li Liu are with the College of Electronic Science and Technology, National University of Defense Technology (NUDT), Changsha 410073, China. Email: pbow16@nudt.edu.cn, lyx\_bible@sina.com, zhoujie\_@nudt.edu.cn, cxd@nudt.edu.cn, everliutianpeng@sina.cn, liuli\_nudt@nudt.edu.cn. 
            \IEEEcompsocthanksitem Xiaogang Yu is with the Beijing Institute of Remote Sensing Information, Beijing 100192, China. Email: 13366351343@189.cn.
        }
    }
		
		
    \markboth{}{Bowen Peng}
		
    \IEEEtitleabstractindextext{
        
    \begin{abstract}
    \justifying
    Joint optical and Synthetic Aperture Radar (SAR) pretraining seeks cross-modal gains over representations learned from either modality alone. This is challenging because the modalities share only partial information, and their correspondence varies across pairs with scene context, acquisition timing, sensor configuration, and spatial resolution. In particular, finer resolution exposes more detailed modality-specific optical appearance and SAR backscatter, further complicating cross-modal supervision. Existing methods encourage cross-modal agreement through representation alignment or cross-modal prediction. Under simplified models of partial, pair-dependent correspondence, we show why a fixed alignment relation can be insufficient and characterize prediction ambiguity arising from target variation unresolved by the source representation. These analyses motivate CoDe-MAE, a masked-autoencoding framework for high-resolution optical-SAR joint pretraining. Its Conditioned Contrastive Learning (CCL) aligns cross-attention-conditioned representations, while Cross-modal Degraded-Target Prediction (CDP) predicts transformed targets to balance conditional predictability and retained supervision. We curate OSPretrain-1M, a 1M-sample optical-SAR corpus, and use CoDe-MAE to pretrain MMATR-X, which achieves leading transfer performance across optical-only, SAR-only, and dual-modal tasks. Controlled experiments on remote-sensing optical-SAR, optical-DEM, and indoor RGB-depth pairs further support these design principles across heterogeneous modality pairs. Code: \url{https://github.com/scenarri/MMATR-X}.
    
    \end{abstract}
    }
    
    \maketitle
    \IEEEdisplaynontitleabstractindextext
    \IEEEpeerreviewmaketitle
		
    \section{Introduction}

    \IEEEPARstart{M}{ulti-sensor} self-supervised learning in Earth observation seeks to exploit relationships across heterogeneous sensing mechanisms so that joint pretraining benefits both single-sensor and multi-sensor downstream applications \cite{hong2026foundation}. Optical and Synthetic Aperture Radar (SAR) imagery represent an important setting of this problem. Optical sensors measure reflected solar radiation and provide rich spectral, textural, and surface-appearance information, but are sensitive to illumination and atmospheric conditions such as cloud cover. SAR actively emits microwave and senses backscatter, capturing dielectric and surface structure, while being independent of solar illumination and less affected by cloud cover, but exhibiting radar-specific effects such as speckle and geometric distortion. 
    
\begin{figure}[tb]
        \centering
        \includegraphics[width=\linewidth]{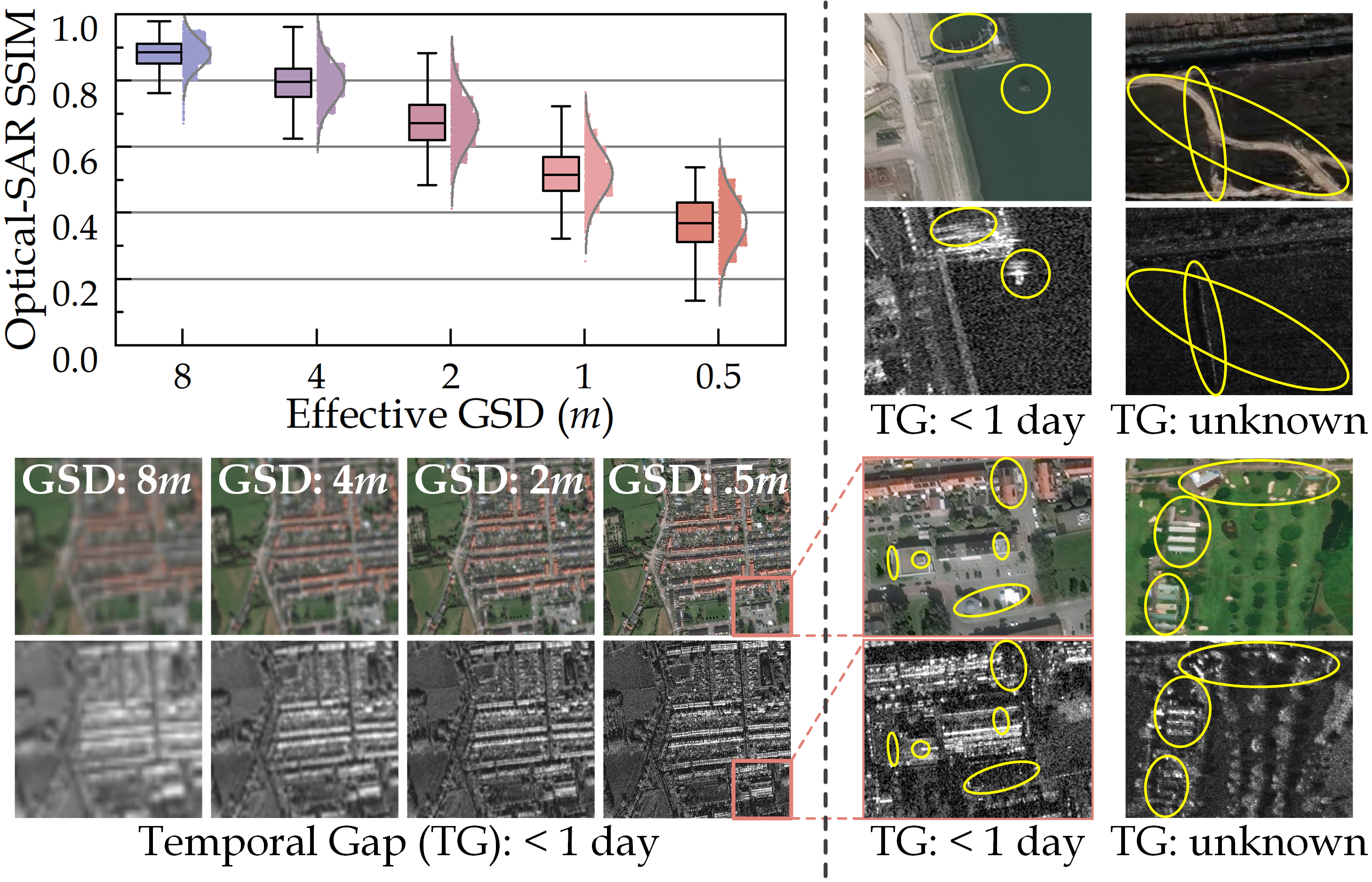}  
        \vspace{-6mm}
        \caption{Optical-SAR correspondence varies across observation pairs and their sensing conditions. \textbf{Left:} The left panel isolates spatial resolution as one important factor affecting correspondence. Starting from the native 0.5-m GSD of MSAW \cite{spacenet6}, we generate coarser effective resolutions using Gaussian-pyramid downsampling following Scale-MAE \cite{reed2023scale}. For pairs acquired within one day, image-level SSIM decreases toward finer GSDs, showing that visual correspondence weakens as finer spatial details are resolved. \textbf{Right:} At the native resolution, representative pairs still exhibit substantially different local discrepancy patterns, including pairs with acquisition gaps below one day and pairs for which the gap is unavailable. Highlighted regions illustrate that the degree and form of local discrepancy vary across individual observations, reflecting the influence of broader pair-specific conditions beyond spatial resolution. Together, these observations motivate treating optical-SAR correspondence as partial and pair-dependent rather than assuming a single fixed cross-modal relation.}
        \label{fig:fig1}
        \vspace{-4mm}
    \end{figure}

    Their correspondence is therefore neither complete nor fixed across observation pairs. The degree and form of correspondence can vary with pair-specific conditions, including scene context, acquisition time gap, sensor configuration, viewing geometry, and spatial resolution, \etc Among these, spatial resolution is particularly important in our setting. High-resolution observations provide substantial value for resolving fine-grained land-cover structures \cite{openearthmapsar,asanet}, recognizing ground objects \cite{li2020object,AID}, and assessing disaster impacts \cite{Chen2025Bright,damageformer}, but finer spatial detail also exposes modality-specific optical appearance and SAR backscatter that may be less apparent at coarser scales. As illustrated in Fig.~\ref{fig:fig1}, the image-level SSIM of optical-SAR pairs acquired within one day decreases toward finer effective Ground Sample Distances (GSDs), indicating weaker visual correspondence as finer spatial details are resolved. Meanwhile, different native-resolution pairs exhibit substantially different local discrepancy patterns even under similar spatial resolution. These observations motivate modeling optical-SAR correspondence as partial and pair-dependent, making a central challenge in high-resolution joint pretraining how to exploit their complementary information while retaining useful modality-specific variation.
 
    Existing optical-SAR joint pretraining methods typically pursue modality gains through two principal forms of cross-modal supervision: representation alignment and cross-modal prediction. Representation-alignment methods encourage paired optical and SAR features to become similar, typically through contrastive learning  \cite{swinssl,croma,yang2026mars,guo2024skysense}. Cross-modal prediction methods instead train the representation of one modality to predict information from its counterpart, commonly through reconstruction objectives \cite{dinomm,msgfm}. Both provide useful supervision: alignment promotes consistency between paired representations, whereas prediction encourages one modality to encode information supported by the other. However, these objectives are generally applied through the same alignment or prediction mechanism across pairs, despite substantial variation in the form and degree of optical-SAR correspondence. Under partial correspondence, such supervision can therefore impose overly restrictive agreement or require prediction of target content that is only weakly determined by the source \cite{Swetha_2023_ICCV,fang2026align}.

    These observations motivate a closer examination of cross-modal supervision under partial and pair-dependent correspondence. For alignment, the central issue is whether a single condition-invariant relation can adequately describe correspondence that changes across observation pairs. Our analysis first formalizes this limitation in a simplified linear setting and then discusses its extension to nonlinear representations, where pair-conditioned mappings can additionally exploit correspondence cues available only from the joint observation. Cross-modal prediction faces a complementary issue: even for observations of the same scene, part of the target variation may remain unresolved from the source representation. We characterize this ambiguity through conditional prediction risk and derive a predictability-retention principle for target transformation. Together, these analyses motivate pair-conditioned alignment and degraded-target prediction.
    
    Following these principles, we introduce Conditioned and Degraded Masked Autoencoder (CoDe-MAE), a masked-autoencoding \cite{mae} framework for high-resolution optical-SAR joint pretraining. It uses modality-specific tokenizers and a shared encoder, with Optical-anchored Knowledge Distillation (OKD) providing supervision from a pretrained optical teacher, allowing both modalities to benefit from the single-modality semantic prior. Conditioned Contrastive Learning (CCL) forms pair-conditioned representations through cross-attention, allowing the alignment relation to adapt to the paired observation. Cross-modal Degraded-Target Prediction (CDP) instead predicts transformed cross-modal targets to balance conditional predictability and retained supervision. A simple grayscale projection, which removes chromatic degrees of freedom while retaining luminance structure, performs well. 
    
\begin{figure}[tb]
        \centering
        \includegraphics[width=\linewidth]{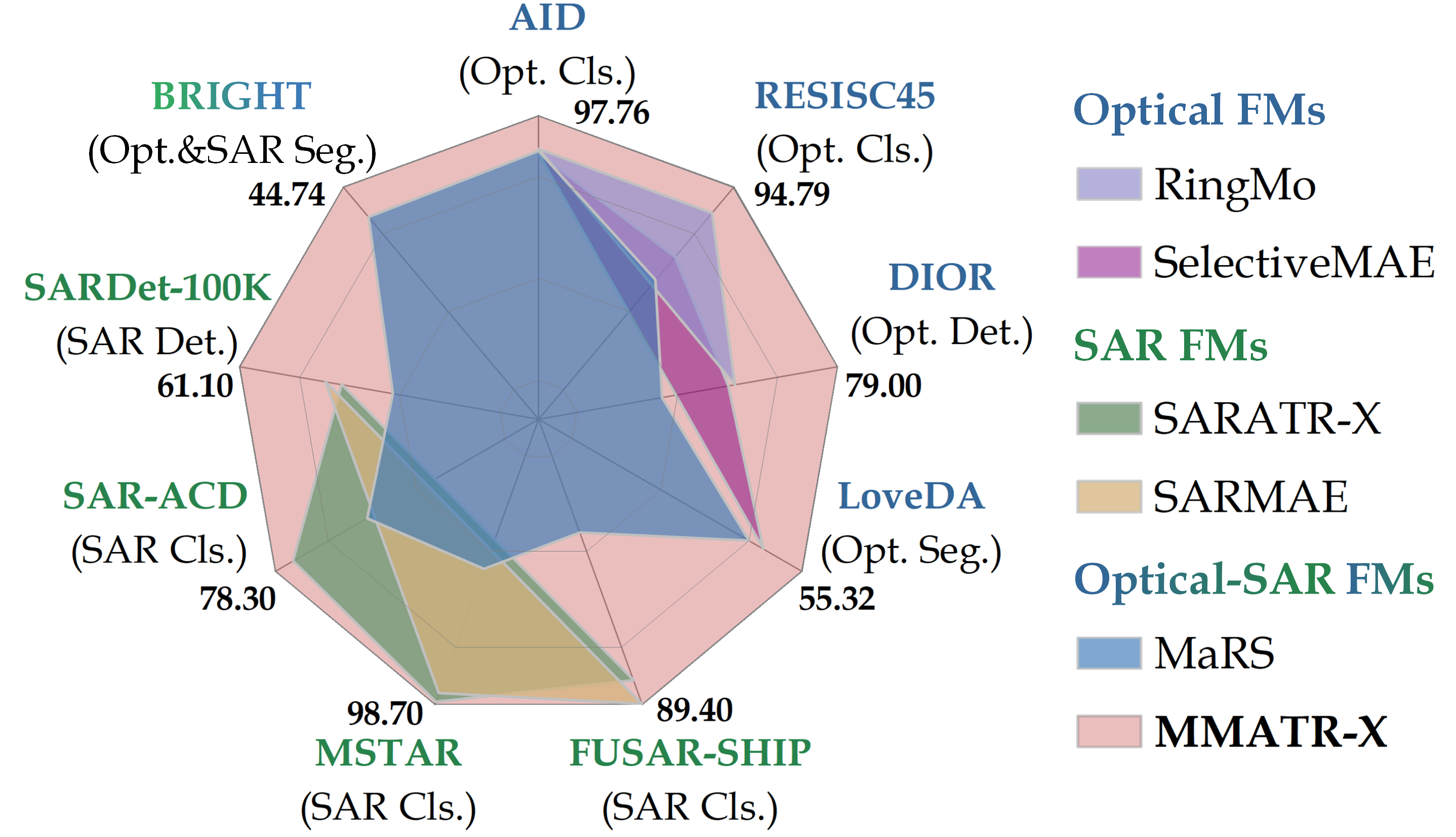}        
        \vspace{-6mm}
        \caption{Comparison of MMATR-X with existing optical, SAR, and optical-SAR foundation models across downstream benchmarks.}
        \label{fig:radar}
        \vspace{-4mm}
    \end{figure}

    We curate OSPretrain-1M, a dataset of over one million paired and unpaired optical and SAR samples, and use it to pretrain MMATR-X using CoDe-MAE. MMATR-X is released as part of the \includegraphics[height=0.7em]{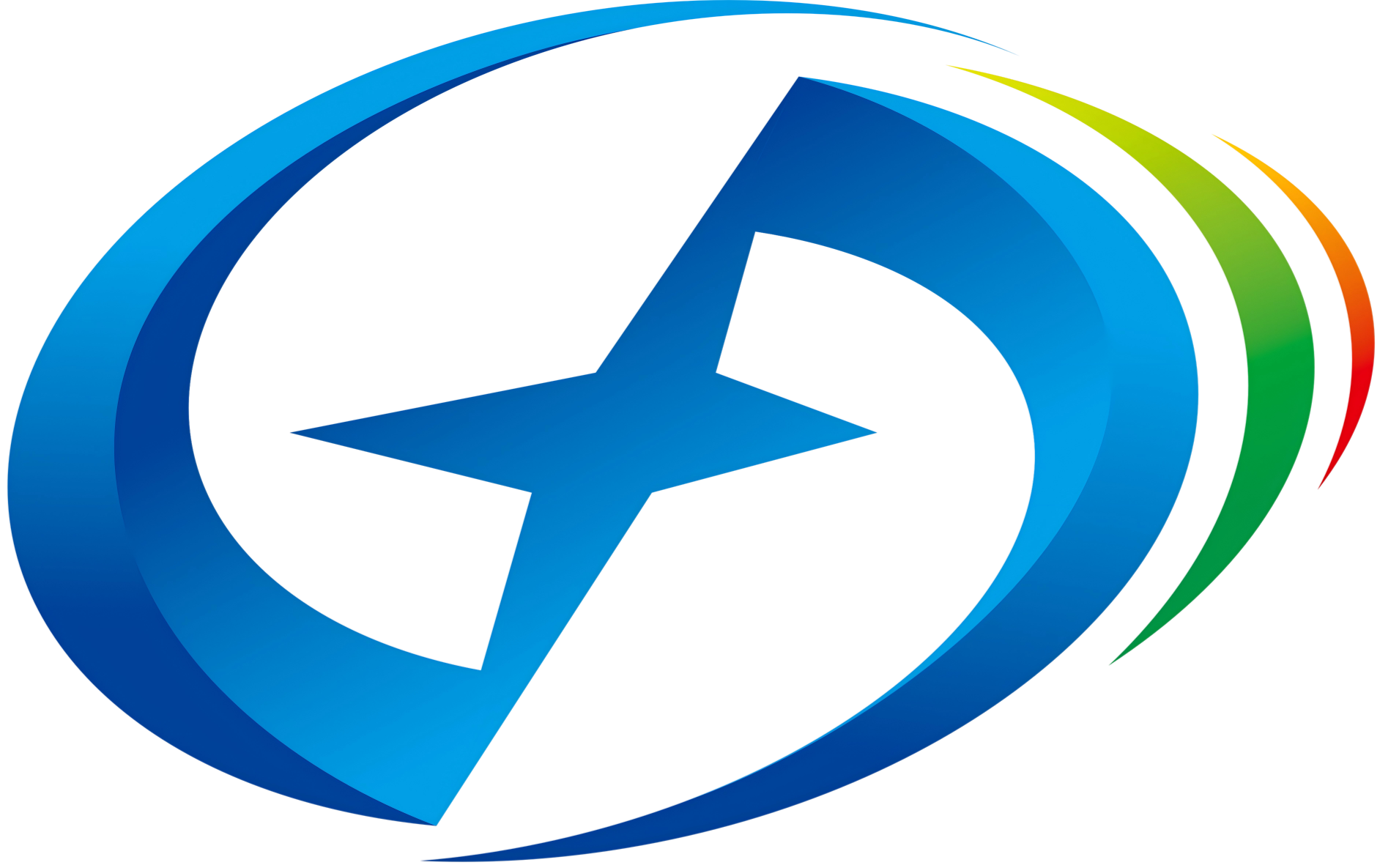}HuiYan model series\cite{saratrx,Mo_2026_CVPR,saratrxv2,huiyanearthsar}, which achieves leading transfer performance across optical-only, SAR-only, and dual-modal downstream tasks spanning classification, segmentation, detection, and change understanding, see Fig.~\ref{fig:radar}. Beyond comparison with existing foundation models, we further conduct controlled experiments to isolate the effect of the pretraining method from differences in backbone, data, and training protocol. Under matched optical-SAR settings, CoDe-MAE compares favorably with representative joint-pretraining methods. We further validate the proposed design on remote-sensing optical-Digital Elevation Model (DEM) and indoor RGB-depth scenarios, where consistent benefits are observed.

    Our main contributions are summarized as follows:
    \begin{enumerate}
    \item We formulate high-resolution optical-SAR joint pretraining under partial and pair-dependent cross-modal correspondence. We analyze the limitation of a fixed alignment relation when correspondence varies across pairs and characterize prediction ambiguity, motivating pair-conditioned alignment and a predictability-retention principle for cross-modal prediction.

    \item We propose CoDe-MAE, which instantiates these principles through CCL for alignment on pair-conditioned representations and CDP for predicting modality-tailored transformed targets. Its modality-shared backbone benefits from pretrained optical-teacher supervision through OKD.
    
    \item We curate OSPretrain-1M, and release MMATR-X pretrained on the full corpus using CoDe-MAE. MMATR-X achieves leading transfer performance across optical-only, SAR-only, and dual-modal downstream tasks. Beyond comparisons with existing foundation models, controlled optical-SAR experiments isolate the effectiveness of CoDe-MAE under matched settings, while optical-DEM and RGB-depth experiments provide additional evidence for the proposed design principles across heterogeneous modality pairs.
    \end{enumerate}

    \section{Related Works}
    
    \subsection{Visual SSL and Knowledge Distillation}

    Visual Self-Supervised Learning (SSL) learns transferable representations from unlabeled images, mainly through contrastive learning, self-distillation, and masked image modeling. Contrastive methods such as SimCLR \cite{simclr} learn invariance by bringing representations of different views of the same image closer while separating unrelated samples. Self-distillation methods such as DINO \cite{dinov1,simeoni2025dinov3} instead match student predictions to targets produced by a teacher network, typically updated from the student without requiring explicit negative pairs. Masked autoendoding methods such as MAE \cite{mae} learn contextual representations by reconstructing masked image regions from the visible content.

    Knowledge distillation differs from self-distillation in that supervision is provided by a separately pretrained teacher \cite{hinton2015distilling}. This paradigm is particularly useful when a strong pretrained model is available for one modality. CoDe-MAE builds on masked autoencoding and uses a frozen optical foundation model to supervise the modality-shared encoder through OKD, allowing both optical and SAR representations to benefit from single-modality distillation.
    
    \subsection{Cross-modal Representation Alignment}
    
    Representation alignment commonly encourages paired observations to be mapped to similar representations. In single-modal SSL, theoretical analyses have shown that alignment tends to emphasize information shared across positive views, while view-specific information may be weakened \cite{tian2020makes,wang2022rethinking}, which can be undesirable when such information remains useful downstream. A similar issue arises in cross-modal learning, where modalities may share only part of their information and at different levels of granularity. CLIP \cite{clip}, for example, aligns images with global textual descriptions, while subsequent vision-language methods introduce richer or denser textual supervision to provide finer-grained correspondence \cite{xiao2025flair,zheng2024dreamlip}.

    Recent multimodal studies have increasingly questioned indiscriminate alignment. Methods that preserve modality-specific structure or explicitly decouple common and unique representations improve learning when modalities contain complementary, non-equivalent information \cite{Swetha_2023_ICCV,decur,hazarika2020misa,qian2026decalign}. Empirical analyses further show that the benefit of alignment depends on modality similarity and the balance between redundant and unique information, and that stronger explicit alignment can either help or hurt under different information structures \cite{tjandrasuwita25a,fang2026align}. These findings motivate moving beyond a single notion of cross-modal agreement. We consider a complementary issue in which the correspondence relation itself varies across observation pairs, motivating pair-conditioned supervision particularly for heterogeneous multi-sensor observations.
    
    \subsection{Optical-SAR Joint Representation Learning}

    Remote-sensing multi-sensor SSL provides a natural testbed for this problem because paired observations from different sensors are intrinsically heterogeneous. Many existing methods are developed on paired medium-resolution observations such as Sentinel-1/2 \cite{sen12ms,ss4eos12}, while recent work has increasingly considered broader downstream transfer and high-resolution optical-SAR imagery.

    For representation alignment, RS-SSL \cite{swinssl} and CROMA \cite{croma} adopt the two modalities as the positive views for contrastive learning, replacing the augmentation-generated views commonly used in single-modal SSL. DeCUR \cite{decur} introduces an explicit decomposition by partitioning the representation dimensions into common and modality-specific subsets. SkySense \cite{guo2024skysense} and MaRS \cite{yang2026mars} adopt dense cross-modal alignment across spatial granularity.

    Another line of work promotes multimodal information integration through joint processing or cross-modal prediction. SatViT \cite{satvit} adopts an early-fusion strategy by concatenating observations from different sensors along the channel dimension and projecting them into a common token representation, which is then processed by a shared backbone. DINO-MM \cite{dinomm} introduces modality dropout and encourages consistency between representations obtained from individual modalities and those obtained from their joint input, thereby supporting both unimodal and multimodal representation learning. msGFM \cite{msgfm} further introduces explicit cross-modal prediction through masked reconstruction, requiring the representation of one modality to recover information from its paired counterpart.
    
    CoDe-MAE shares the general goal of exploiting paired optical-SAR observations, but focuses on adapting cross-modal supervision to partial and pair-dependent correspondence. CCL performs alignment on pair-conditioned representations, allowing the alignment relation to adapt to the paired observation. CDP instead predicts transformed cross-modal targets that reduce conditionally ambiguous variation while retaining useful supervision.

    \section{Cross-modal Supervision under Partial and Pair-Dependent Correspondence}
    \label{sec:theory}
    
    We analyze cross-modal alignment and prediction for high-resolution optical-SAR joint pretraining under partial and pair-dependent correspondence. We use tractable population-level abstractions to study where cross-modal alignment is applied and what target information is predicted. Prior analyses of contrastive learning have shown that shared information plays a central role in cross-view representation learning, while non-shared information can remain relevant to downstream tasks \cite{tian2020makes,wang2022rethinking}. Recent multimodal studies further show that the benefit of cross-modal alignment depends on the relation between shared and modality-specific information \cite{almudevar2025aligning,tjandrasuwita25a,fang2026align}. We present the core analysis here and defer detailed proofs and illustrative examples to the supplementary material.
    
    \subsection{Problem Formulation}
    
    We use linear latent-factor abstractions of paired optical and SAR observations and their representations. Throughout the analysis, encoders are held fixed, and all random vectors have finite second moments. At the
    observation level, let
    \begin{equation}
        \mathbf x_o = B_o^{(c)}\mathbf s+\mathbf u_o,
        \qquad
        \mathbf x_s = B_s^{(c)}\mathbf s+\mathbf u_s,
        \label{eq:partial_observation}
    \end{equation}
    where $\mathbf s$ denotes latent factors shared by the paired observations, $c$ denotes the correspondence condition of the current pair, while $\mathbf u_o$ and $\mathbf u_s$ collect modality-specific variation. The mappings $B_o^{(c)}$ and $B_s^{(c)}$ describe how the shared scene factors are manifested under the sensing and acquisition conditions of that pair. The condition $c$ can abstract pair-specific factors such as scene context, acquisition time gap, sensor configuration, viewing geometry, and spatial scale, \etc Consequently, even when two observations depict the same scene, their shared information need not appear in the same form, and each modality can contain variation that is not uniquely determined by its counterpart.
    
    Let $\mathbf z_m=f_m(\mathbf x_m), m\in\{o,s\}$ denote the corresponding encoder representation. For analyzing representation alignment, we separately posit the representation-level abstraction
    \begin{equation}
        \mathbf z_o = A_o^{(c)}\mathbf s+\mathbf v_o,
        \qquad \mathbf z_s = A_s^{(c)}\mathbf s+\mathbf v_s,
        \label{eq:partial_representation}
    \end{equation}
    where $\mathbf v_o$ and $\mathbf v_s$ denote non-shared information retained in the representation. Eq.~\eqref{eq:partial_representation} is an analytical abstraction, not a consequence of applying an arbitrary nonlinear encoder to Eq.~\eqref{eq:partial_observation}. These models describe two aspects of partial and pair-dependent correspondence: shared factors can appear in different forms across pairs, and one modality need not fully determine the other.
    
    \subsection{Representation Alignment}\label{sec:analysis_alignment}
    
    We first isolate the positive-pair pressure induced by representation
    alignment. Consider the orthogonal special case of
    Eq.~\eqref{eq:partial_representation},
    \begin{equation}
        \mathbf z_o=\mathbf s+\mathbf v_o,
        \qquad
        \mathbf z_s=\mathbf s+\mathbf v_s,
    \end{equation}
    with $\mathbf s$, $\mathbf v_o$, and $\mathbf v_s$ mutually orthogonal and $\|\mathbf z_o\|_2,\|\mathbf z_s\|_2>0$. Their cosine similarity is
    \begin{equation}
    \operatorname{sim}(\mathbf z_o,\mathbf z_s) = \frac{\|\mathbf s\|_2^2}
    {\sqrt{ (\|\mathbf s\|_2^2+\|\mathbf v_o\|_2^2)(\|\mathbf s\|_2^2+\|\mathbf v_s\|_2^2)}}.
    \label{eq:shared_private_cos}
    \end{equation}
    For fixed $\mathbf s$, Eq.~\eqref{eq:shared_private_cos} is non-increasing with the energy of either non-shared component, and strictly decreasing when $\|\mathbf s\|_2>0$. This result only isolates the positive-pair attraction term; the full contrastive objective additionally includes negative-pair discrimination and other objectives that can favor instance-specific information. Nevertheless, if modality-specific components contain downstream-relevant information, attenuating them to increase cross-modal agreement can be undesirable \cite{wang2022rethinking}.
    
    Modern contrastive methods commonly introduce projection heads to separate the alignment space from the backbone representation \cite{simclr,clip}. These heads allow the alignment representation to differ from the base representation, but are typically formed independently for each modality: $g_o(\mathbf z_o)$ does not depend on $\mathbf z_s$, and vice versa. Under Eq.~\eqref{eq:partial_representation}, a fixed projection must therefore accommodate different pair-dependent correspondence structures through the same mapping.
    
    To examine fixed linear alignment, for each correspondence condition $c$, define the conditionally centered cross-modal covariance
    \begin{equation}
        M_c = \operatorname{Cov}(\mathbf z_o,\mathbf z_s\mid c).
        \label{eq:conditional_cross_cov}
    \end{equation}
    Spectral analyses have related linear multimodal contrastive optimization to a parameter-dependent contrastive cross-covariance incorporating both positive and negative pairs \cite{nakada2023understanding}. Inspired by this perspective, we consider a simpler positive-pair dot-product criterion to isolate the effect of varying correspondence conditions.
    
    \begin{proposition}[Fixed vs.\ condition-dependent linear alignment]
    \label{prop:conditional_alignment}
    Let $c$ take values in a finite set $\mathcal C$ with probabilities $p_c$. Assume $\mathbb E[\mathbf z_m\mid c]=0$ for $m\in\{o,s\}$, and let $1\le r\le\min(d_o,d_s)$, where $d_m$ is the dimension of $\mathbf z_m$.
    For maps $W_m^\top\mathbf z_m$ with $W_m^\top W_m=I_r$, the maximum expected positive-pair dot product achieved by one fixed pair of maps and by oracle maps with access to $c$ satisfies
    \begin{equation}
        \mathcal A_{\mathrm{fix}}^{(r)} = \left\|\mathbb E_c[M_c]\right\|_{(r)}
        \le \mathbb E_c\!\left[\|M_c\|_{(r)}\right] = \mathcal A_{\mathrm{cond}}^{(r)},
        \label{eq:conditional_alignment_bound}
    \end{equation}
    where $\|\cdot\|_{(r)}$ denotes the sum of the largest $r$ singular values.
    \end{proposition}
    
    The inequality follows from the variational characterization and convexity of the Ky-Fan norm. It is strict when the conditions with $p_c>0$ do not admit a common maximizing pair of left and right projections, for example when their cross-covariances partially cancel. Thus, at fixed encoder representations, access to the correspondence condition can increase the attainable positive-pair alignment in this linear surrogate. This is an expressivity comparison; it does not order the full normalized contrastive losses or their encoder gradients. The proof, equality condition, illustrative example, and relation to InfoNCE are provided in the supplementary material.
    
    A nonlinear counterpart-independent projector can exploit correspondence cues available in its own modality. Supplementary Section~3.3 considers a special case in which the correspondence condition cannot be identified from either modality alone but can be identified from their pair. In this setting, pair-conditioned mappings admit aligned solutions without the correspondence-invariance constraint on projection outputs required by counterpart-independent mappings. This existence result concerns the alignment space, rather than information loss in the underlying encoder. Together, these results motivate pair-conditioned alignment by giving the alignment function access to its paired representation when correspondence cues are unavailable from either modality alone.
    
    \subsection{Cross-modal Prediction}\label{sec:analysiscdp}
    
    The observation-level model also describes ambiguity in cross-modal prediction. Consider SAR-to-optical prediction, where a fixed source representation $\mathbf z_s$ is used to predict an optical target $\mathbf x_o$. In masked prediction, the target can be a patch sampled from the masked locations, with the source tokens, mask, and target position included in the conditioning information. Under squared error, the Bayes-optimal predictor is
    \begin{equation}
        f^*(\mathbf z_s) = \mathbb E[\mathbf x_o\mid\mathbf z_s],
    \end{equation}
    with minimum achievable risk
    \begin{equation}
        R_{\mathrm{full}}^* = \mathbb E\!\left[ \operatorname{tr} \operatorname{Cov}(\mathbf x_o\mid\mathbf z_s) \right].
        \label{eq:crossmodal_bayes}
    \end{equation}
    Whenever this conditional variance is nonzero, full-target prediction involves variation that cannot be fully determined from the given source representation.
    
    Eq.~\eqref{eq:partial_observation} provides an interpretation of this ambiguity. Within a fixed correspondence condition $c$, assume that $\mathbf u_o$ has zero mean and is independent of $(\mathbf s,\mathbf z_s)$ under the conditional distribution. The minimum risk in this condition is
    \begin{equation}
        R_c^* = \operatorname{tr} \left( B_o^{(c)}\overline{\Sigma}_{\mathbf s\mid\mathbf z_s,c}(B_o^{(c)})^\top \right) + \operatorname{tr}(\Sigma_{\mathbf u_o\mid c}),
        \label{eq:full_target_risk}
    \end{equation}
    where
    $\overline{\Sigma}_{\mathbf s\mid\mathbf z_s,c} = \mathbb E[\operatorname{Cov}(\mathbf s\mid\mathbf z_s,c)\mid c]$ and $\Sigma_{\mathbf u_o\mid c}=\operatorname{Cov}(\mathbf u_o\mid c)$.
    The first term reflects shared factors that remain unresolved from the source representation, whereas the second corresponds to optical-specific variation that remains unpredictable under this analytical model. Variation in the correspondence condition can introduce additional uncertainty; all such sources are captured by the general conditional covariance in Eq.~\eqref{eq:crossmodal_bayes}.
    
    This observation motivates separating target variation according to its
    predictability from the source. By the law of total covariance,
    \begin{equation}
        \Sigma_{\mathbf x_o} = \underbrace{ \operatorname{Cov}\left(\mathbb E[\mathbf x_o\mid\mathbf z_s] \right)}_{\Sigma_{\mathrm{pred}}}+\underbrace{\mathbb E\left[\operatorname{Cov}(\mathbf x_o\mid\mathbf z_s)\right]}_{\Sigma_{\mathrm{amb}}},
        \label{eq:pred_amb_decomposition}
    \end{equation}
    where $\Sigma_{\mathrm{pred}}$ characterizes target variation explained by the conditional mean under squared error, and $\Sigma_{\mathrm{amb}}$ the remaining conditional ambiguity.
    
    Consider a linear degraded target $\widetilde{\mathbf x}_o=P\mathbf x_o$. Its irreducible squared prediction
    risk is
    \begin{equation}
        R_P^* = \operatorname{tr} \left(P\Sigma_{\mathrm{amb}}P^\top\right).
    \end{equation}
    Reducing this quantity alone is insufficient, since a transformation that removes all target variation would also remove useful supervision. To illustrate this trade-off at a fixed scale and output dimension, consider
    $P\in\mathbb R^{r\times D_o}$ with orthonormal rows, where $D_o$ is the target dimension and $1\le r\le D_o$. An idealized predictability-retention criterion is
    \begin{equation}
        \min_P\:\operatorname{tr}
        \left(P\Sigma_{\mathrm{amb}}P^\top\right),\:
        \mathrm{s.t.}\:\operatorname{tr}\left(P\Sigma_{\mathrm{pred}}P^\top\right)
        \ge \kappa,\:PP^\top=I_r.
    \label{eq:predictability_retention}
    \end{equation}
    Here, $\kappa>0$ specifies the minimum retained variation explained by the source-conditioned mean and is chosen within the feasible range derived in the supplement. The orthonormal-row constraint fixes the scale of this illustrative linear family. The criterion expresses a trade-off between conditional predictability and retained supervision; the choice of a practical degraded-target transformation depends on the modality pair and is assessed empirically.
    
    \subsection{Design Implications}
    
    The above analyses suggest two complementary principles for cross-modal supervision under partial and pair-dependent correspondence. First, pair-conditioned alignment can exploit correspondence cues unavailable to a counterpart-independent mapping. Second, degraded-target prediction should balance conditional ambiguity against retention of target variation that remains predictably supported by the source. These principles motivate CCL and CDP, respectively. The analyses characterize idealized alignment capacity and prediction risk, while their implications for learned representations are evaluated empirically.

    \section{CoDe-MAE Pretraining Framework}\label{methods}

    \begin{figure*}[htbp]
        \centering
        \includegraphics[width=\linewidth]{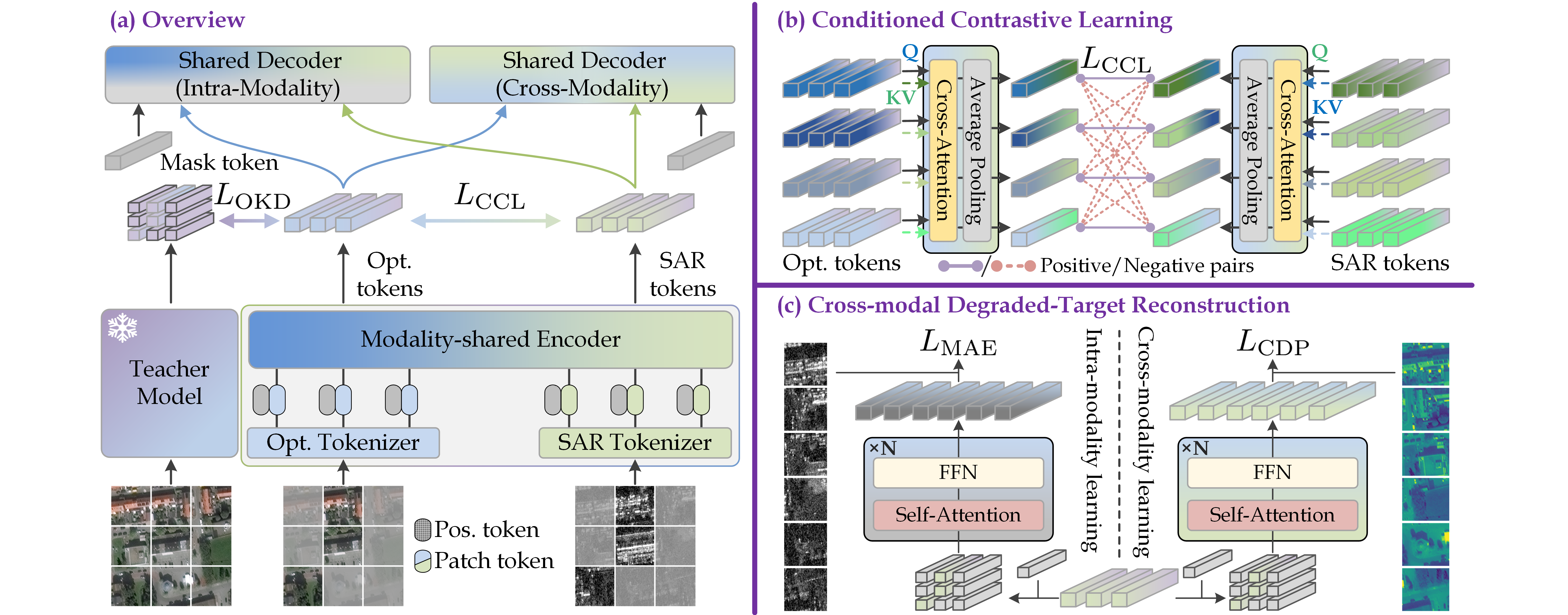}
        \caption{{Overview of the CoDe-MAE pretraining framework.} ({a}) Optical-anchored Knowledge Distillation (OKD) regularizes the modality-shared encoder using representations from a frozen optical teacher (Section~\ref{sec:okd}). ({b}) Conditioned Contrastive Learning (CCL) first introduces information from the paired modality through cross-attention and then applies contrastive supervision to the resulting conditioned representations, rather than directly aligning the independently encoded optical and SAR features (Section~\ref{sec:ccl}). ({c}) Cross-modal Degraded-Target Prediction (CDP) predicts transformed cross-modal targets motivated by a trade-off between conditional ambiguity and retained predictable variation (Section~\ref{sec:cdp}). After pre-training, only the modality-specific tokenizers and the shared encoder are retained for downstream task adaptation.}
       
        \label{fig:method}
    \end{figure*}
    
    \subsection{Overview and Base Masked Autoencoder}

    As shown in Fig.~\ref{fig:method}, CoDe-MAE combines a modality-shared masked-autoencoding backbone with three auxiliary pretraining objectives: OKD, CCL, and CDP. Given optical and SAR observations, modality-specific tokenizers map the inputs into a common token space, after which a shared encoder extracts their representations. CCL implements pair-conditioned alignment, while CDP introduces degraded-target cross-modal prediction for registered pairs; OKD provides optical-teacher supervision through the shared encoder. All auxiliary modules are discarded after pretraining, leaving the modality-specific tokenizers and shared encoder for downstream adaptation.

    \textbf{Masked autoencoding \cite{mae}.}
    An input image $\mathbf{x}_{m}\in\mathbb{R}^{C\times H\times W}$ is partitioned into $M$ patches $\mathbf{P}_{m}$. A binary mask $\mathbf{m}\in\{0,1\}^{M}$ is randomly sampled, where $\mathbf{m}_i=1$ denotes a masked patch. Only the $M_{\mathrm{vis}}$ visible patches are encoded into latent representations $\mathbf{z}_{m}\in\mathbb{R}^{M_{\mathrm{vis}}\times D}$,
    while a Transformer decoder reconstructs the masked input patches using the encoded visible tokens and learnable mask tokens. For paired optical-SAR observations, we use the same spatial mask for both modalities so that their visible and masked regions remain spatially corresponding. Unless otherwise specified, the masking ratio is 75\%.
    
    The modality-wise reconstruction objective is
    \begin{equation}
        L_{\mathrm{MAE}}
        =
        \frac{\sum_{i:\mathbf{m}_i=1}
        \left\|\mathbf{R}_{o,i}-\mathbf{P}_{o,i}\right\|_2^2}{M_{\mathrm{mask}}}
        +
        \frac{\sum_{i:\mathbf{m}_i=1}
        \left\|\mathbf{R}_{s,i}-\mathbf{P}_{s,i}\right\|_2^2}{M_{\mathrm{mask}}},
    \end{equation}
    where $\mathbf{R}_{o}$ and $\mathbf{R}_{s}$ denote the reconstructed optical and SAR patches, respectively, and $M_{\mathrm{mask}}$ is the number of masked patches.

    \subsection{Optical-anchored Knowledge Distillation}\label{sec:okd}

    Optical imagery provides rich semantic appearance cues and benefits from mature pretrained visual foundation models, whereas SAR observations follow a substantially different electromagnetic sensing mechanism and exhibit radar-specific effects such as speckle noise. We therefore use a pretrained optical model to provide semantic supervision during joint pretraining, without directly forcing optical and SAR representations to match.
    
    Specifically, we employ the frozen DINOv3-Sat model \cite{simeoni2025dinov3} as an optical teacher $f_{\mathrm{teacher}}$. The optical and SAR inputs are processed by the same student encoder, while knowledge distillation is applied only to the optical branch. For each visible optical patch, the corresponding student representation is encouraged to match the teacher representation using an $\ell_1$ objective:
    \begin{equation}
        L_{\mathrm{OKD}}=\frac{1}{M_{\mathrm{vis}}}\sum_{i:\mathbf{m}_i=0}\left\|f_{\mathrm{teacher}}(\mathbf{x}_{o})_i-\mathbf{z}_{o,i}\right\|_1 .
    \end{equation}

    Because the student encoder is shared across modalities, optical-only distillation updates the same parameters that are subsequently used to encode SAR observations. This provides a direct pathway for the optical teacher signal to influence the SAR representation without explicitly forcing paired optical and SAR features to match. OKD therefore serves as a semantic regularizer for the shared encoder, upon which the following cross-modal objectives are introduced.
    
 
    \subsection{Conditioned Contrastive Learning}\label{sec:ccl}

    Following the analysis in Section~\ref{sec:theory}, CCL instantiates pair-conditioned alignment by allowing the representation presented to the contrastive objective to depend on its paired counterpart, rather than using a counterpart-independent projection.
    
    Given independently encoded optical and SAR visible-token sequences $\mathbf{z}_{o}$ and $\mathbf{z}_{s}$, we construct two direction-specific conditioned representations:
    \begin{equation}
    \begin{aligned}
    \mathbf{z}^{cd}_{o\leftarrow s} &= \mathbf{z}_{o} + \operatorname{CA}_{o}(\mathbf{z}_{o},\mathbf{z}_{s}),\\
    \mathbf{z}^{cd}_{s\leftarrow o} &= \mathbf{z}_{s} + \operatorname{CA}_{s}(\mathbf{z}_{s},\mathbf{z}_{o}),
    \end{aligned}
    \label{eq:ca}
    \end{equation}
    where $\operatorname{CA}_{o}$ and $\operatorname{CA}_{s}$ are independently parameterized multi-head cross-attention modules \cite{vaswani2017attention}. Each module uses the first modality as queries and the counterpart modality as keys and values, with pre-normalization. The residual formulation retains the independently encoded tokens while augmenting them with information retrieved from the paired modality.
    
    We then mean-pool the conditioned tokens and apply $\ell_2$ normalization to obtain the global representations $\bar{\mathbf{z}}^{cd}_{o\leftarrow s}$ and $\bar{\mathbf{z}}^{cd}_{s\leftarrow o}$. A symmetric contrastive objective is applied:
    \begin{multline}
    L_{\mathrm{CCL}} = -\frac{1}{2N} \sum_{i=1}^{N} \left[\log \frac{\exp(\langle\bar{\mathbf{z}}^{cd}_{o\leftarrow s,i},\bar{\mathbf{z}}^{cd}_{s\leftarrow o,i}\rangle/\tau)}{\sum_{j=1}^{N}\exp(\langle\bar{\mathbf{z}}^{cd}_{o\leftarrow s,i},\bar{\mathbf{z}}^{cd}_{s\leftarrow o,j}\rangle/\tau)}\right.\\
    \left. +\log\frac{\exp(\langle\bar{\mathbf{z}}^{cd}_{s\leftarrow o,i},\bar{\mathbf{z}}^{cd}_{o\leftarrow s,i}\rangle/\tau)}{\sum_{j=1}^{N}\exp(\langle\bar{\mathbf{z}}^{cd}_{s\leftarrow o,i},\bar{\mathbf{z}}^{cd}_{o\leftarrow s,j}\rangle/\tau)}
    \right],
    \end{multline}
    where $N$ is the batch size and $\tau=0.07$. 
    
    CCL instantiates the pair-conditioned alignment principle on which cross-modal alignment is performed: contrastive supervision is applied after pair conditioning rather than directly to the independently encoded optical and SAR representations. 

    \subsection{Cross-modal Degraded-Target Prediction}\label{sec:cdp}

    Following the predictability-retention principle, CDP transforms the cross-modal prediction target to reduce conditionally ambiguous variation while retaining useful supervision.
    
    Specifically, a target-side operator $\operatorname{Deg}(\cdot)$ is applied before cross-modal prediction. Let $\mathbf{R}^{cdp}_{o\rightarrow s}$ denote the SAR prediction from the optical representation and $\mathbf{R}^{cdp}_{s\rightarrow o}$ the optical prediction from the SAR representation. Using a modality-shared auxiliary decoder, CDP is optimized as    
    \begin{multline}
    L_{\mathrm{CDP}} = \frac{1}{M_{\mathrm{mask}}} \sum_{i:\mathbf{m}_i=1} \left\|\mathbf{R}^{cdp}_{o\rightarrow s,i} - \operatorname{Deg}(\mathbf{P}_{s,i})\right\|_2^2
    \\
    + \frac{1}{M_{\mathrm{mask}}} \sum_{i:\mathbf{m}_i=1}
    \left\| \mathbf{R}^{cdp}_{s\rightarrow o,i} - \operatorname{Deg}(\mathbf{P}_{o,i}) \right\|_2^2.
    \end{multline}

    CDP modifies the prediction target to focus supervision on counterpart information that is more reliably supported by the source modality. For optical-SAR pretraining, $\operatorname{Deg}$ converts the optical target to grayscale using the fixed luminance projection $0.299R+0.587G+0.114B$, removing chromatic degrees of freedom while retaining luminance-based spatial structure at the original resolution. Since we use single-channel SAR imagery, $\operatorname{Deg}$ is the identity. Suitable target transformations can differ across modality pairs, as examined in Section~\ref{sec:ablationcdp}.

    \subsection{Overall Objective and Data Requirement}\label{datausage}

    The four pretraining objectives operate on different levels of data requirement.
    For a training sample, the overall objective is written as
    \begin{equation}    L_{\mathrm{Total}}=L_{\mathrm{MAE}}+\delta_{\mathrm{opt}}L_{\mathrm{OKD}}+\delta_{\mathrm{pair}}\left(L_{\mathrm{CCL}}+L_{\mathrm{CDP}}\right),
    \end{equation}
    where $\delta_{\mathrm{opt}},\delta_{\mathrm{pair}}\in\{0,1\}$ indicate the availability of an optical observation and a spatially corresponding optical-SAR pair, respectively. Masked reconstruction is applied to every available modality, while OKD requires only an optical observation. CCL and CDP require paired optical-SAR observations and are therefore activated only for the paired subset.

    This flexible objective allows heterogeneous pretraining data to be organized according to the level of cross-modal correspondence they provide: paired samples support CCL and CDP, while unpaired observations contribute through modality-wise reconstruction and optical-teacher supervision. This enables CoDe-MAE to make use of a substantially broader pretraining corpus than would be possible with strictly paired data alone.

    \begin{table*}[tb]
    \centering
    \caption{Composition and provenance of OSPretrain-1M. $^{\dag}$The test split is excluded from pretraining. GEE denotes multi-source imagery accessed through Google Earth Engine. Geographic coverage is reported according to the information released by the source datasets.}
    \label{tab:ospretrain1m}
    \vspace{-2mm}
    \resizebox{\linewidth}{!}{
    \setlength{\tabcolsep}{3pt}
    \begin{tabular}{l|cccccl}
        \toprule
        {\textbf{Dataset}} & \textbf{Year} & \textbf{Modality} & \textbf{Source}  & \textbf{$\#$Patches} & \textbf{Resolution} & \textbf{Coverage/Description}\\   \hline
        \multicolumn{2}{l}{\textit{Paired subset}} \\ \hline
        MSAW \cite{spacenet6} & 2020 & O\&S & {Worldview-2, Airborne SAR}  & 64,084 & .5/.5  &   Rotterdam\\
        QXS-SAROPT	\cite{qxssarport} & 2021 & O\&S & GEE/GF-3 & 40,000 &  1/1 &  San Diego, Shanghai, and Qingdao \\ 
        DDHR-DY	\cite{ddhrnet}& 2022 & O\&S & GF-2/GF-3 & 15,662 & 1/1 &  Xi'an and Dongying \\ 
        WHU-OPT-SAR$^{\dag}$	\cite{whuoptsar} & 2022 & O\&S & GF-1/GF-3 & 47,040 & 2/5 &  Hubei Province \\  
        DFC23-T1/T2$^{\dag}$ \cite{persello20232023} & 2023 & O\&S & SuperView-1, GF-2/GF-3 & 65,800 & .5-1/1 & 17 cities over 6 continents \\ 
        OGSOD-2 \cite{ogsod2} & 2025 & O\&S & GEE/GF-3 & 32,570 & 3/10 &  - \\  
        GFGE-SO \cite{gfgedataset} & 2025 & O\&S & GEE/GF-3 & 60,000 & 5/5 & 7 cities in China
        \\
        OEM-SAR \cite{openearthmapsar} & 2025 & O\&S & {NAIP, IGN \& GSI/Umbra} & 138,578 & .15-.5/$<$1 & 35 cities in Japen, France, and the USA \\ 
        EarthMiss \cite{zhou2026remote} & 2025 & O\&S & Worldview-2/Hisea-1, Cappella & 44,008 & .6/.35 & 13 cities distributed across five continents\\
        OSdataset 2.0 \cite{osdataset2} & 2025 & O\&S & GEE/GF-3 & 30,424 & .5/.5 & 14 countries worldwide\\
        \hline
        Subtotal &  &  &  & 538,166 \\  \hline
        \multicolumn{2}{l}{\textit{Unpaired subset}} \\ \hline
        RSD46-WHU$^{\dag}$ \cite{long2017accurate} & 2017 & O & GEE, Tianditu  & 109,157 & .5-2 & - \\
        FAIR1M$^{\dag}$ \cite{sun2022fair1m} & 2022 & O & GF, GEE & 149,881 & .3-.8 & more than 100 civil airports, harbors, and cities worldwide \\
        FAIR-CSAR$^{\dag}$ \cite{wu2024fair} & 2024 & S & GF-3 & 33,214 & 1, 5 & airports, refineries, ports, and river channels worldwide  \\
        SARDet-180K$^{\dag}$ \cite{saratrx} & 2025 & S & Airborne SAR, GF-3, RadarSat-2, \etc & 157,733 & - & a composite pretraining set that integrates 14 SAR datasets \\
        ATRNet-STAR$^{\dag}$ \cite{liu2025atrnet} & 2025 & S & Aerial SAR & 68,091 & .15 & - \\ \hline
        Subtotal &  &  &  & 518,076 \\ 
        \bottomrule
    \end{tabular}}
    \vspace{-4mm}
    \end{table*}

    \subsection{OSPretrain-1M Dataset}

    To exploit this flexibility at scale, we curate OSPretrain-1M from 15 publicly available datasets, combining both paired and unpaired optical and SAR observations. This is particularly useful for high-resolution optical-SAR pretraining, where large-scale reliably paired data remain relatively limited and fragmented across existing datasets. The resulting dataset comprises approximately 1.06M training patches (Table~\ref{tab:ospretrain1m}), including 538K patches forming approximately 269K geo-registered optical-SAR pairs and 518K unpaired optical or SAR patches, spanning multiple sensors, spatial resolutions, geographic regions, and scene types. 

    The paired subset consists of observations for which reliable spatial correspondence can be established and is used by CCL and CDP. The unpaired subset broadens the modality-specific content and sensor distributions, and also includes target-centric imagery such as aircraft and ships for which reliable cross-modal instance correspondence is unavailable or difficult to establish. These samples are retained as unpaired observations and used only with objectives that do not require cross-modal correspondence.

    \begin{figure}[tb]
        \centering
        \includegraphics[width=\linewidth]{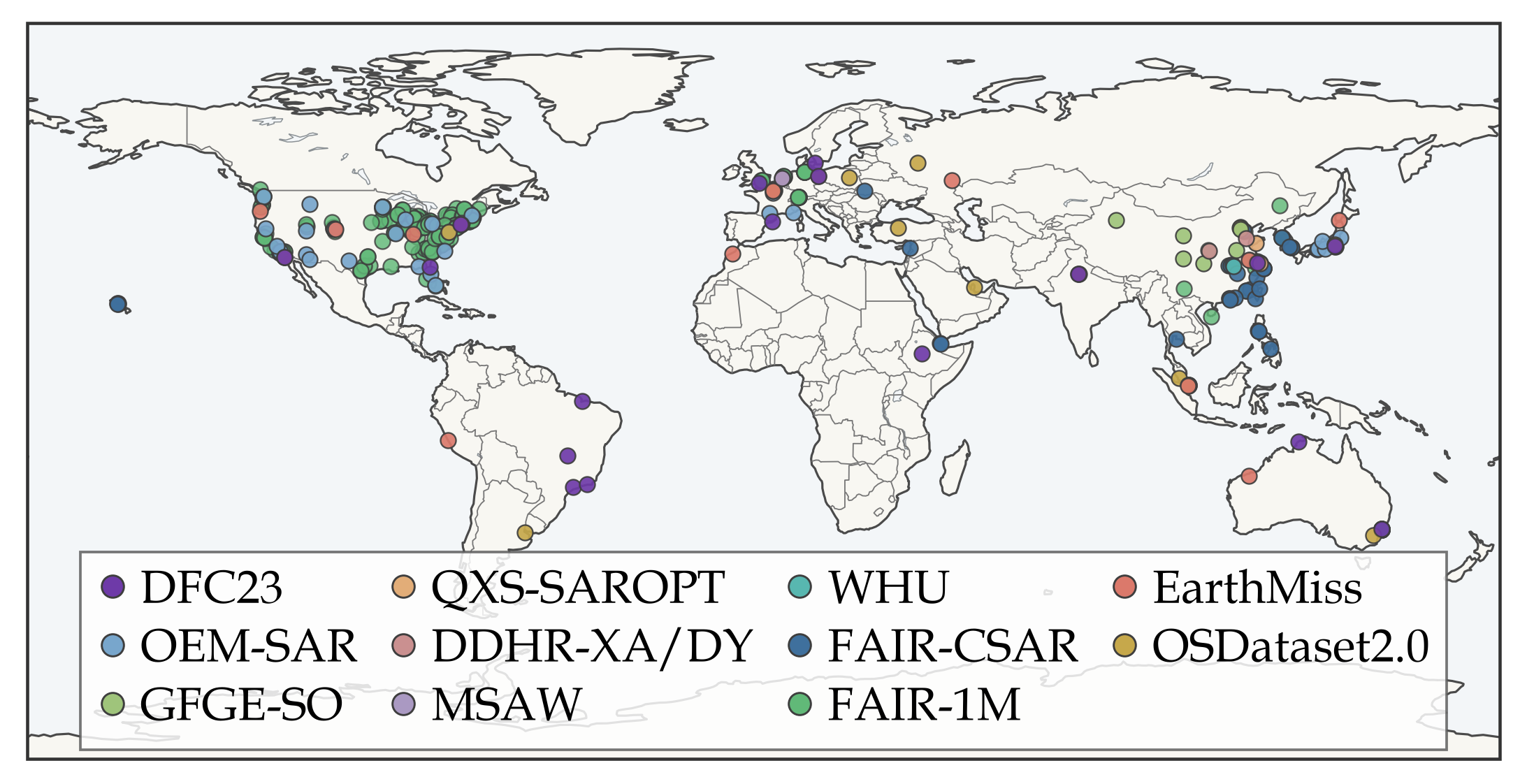}
    \vspace{-8mm}
        \caption{Reported city-level geographic provenance of OSPretrain-1M. Markers denote locations that can be identified from publicly released information, yet do not represent complete coverage of the corresponding administrative regions. Datasets without sufficiently specific geographic information are not assigned.}
    \label{fig:geo_distri}
    \vspace{-4mm}
    \end{figure}
    
    We summarize the geographic provenance that can be identified for the constituent datasets. Fig.~\ref{fig:geo_distri} reports city-level locations when acquisition regions can be determined from public metadata or dataset descriptions. Most source datasets provide only city- or region-level provenance rather than image-level coordinates; therefore, these markers indicate reported acquisition locations rather than complete geographic footprints. The image-level overlap audit used in the downstream comparisons is described separately in Section~\ref{sec:dataoverlapautid}.

    \section{Experiments}

    We first pre-train CoDe-MAE on OSPretrain-1M for 800 epochs to obtain the MMATR-X foundation model and compare it with existing remote sensing foundation models. We then conduct fully controlled experiments on tasks with different modality configurations to evaluate the proposed method under matched settings. Finally, we present ablation studies and mechanism analyses to examine the contributions and behavior of the main designs.

    \subsection{Implementation Details}

    To bridge pretraining and dense prediction tasks, CoDe-MAE utilizes the iTPN framework \cite{itpn} equipped with an ImageNet-1K pre-trained HiViT-Base/16 encoder (79M parameters) weights \cite{hivit}. For the OKD teacher, we employ a frozen DINOv3-Sat (ViT-L) model for remote sensing optical-SAR and optical-DEM and DINOv3-Lvd (ViT-L) for indoor RGB-depth \cite{simeoni2025dinov3}. Models are optimized using AdamW with a total batch size of 2,048 and a base learning rate of $1.5\times 10^{-4}$, scheduled via cosine annealing with a 15-epoch warmup. Pretraining is distributed across eight NVIDIA RTX 4090D GPUs, spanning 800 epochs for MMATR-X and, unless otherwise specified, 100/200 epochs for controlled comparisons and ablation studies. In all tables, the best and second-best results are highlighted in \tone{red} and \ttwo{blue}, respectively. Additional implementation details and qualitative results are provided in our supplementary.

    \subsection{Comparison with Existing Foundation Models}\label{comp_sota}

    We compare MMATR-X with existing optical, SAR, and multimodal remote-sensing foundation models. Since these models differ substantially in architecture, pretraining data, and pretraining recipes, and are typically evaluated under different downstream protocols, we compare them within the evaluation settings most relevant to each model family rather than enforcing a single common benchmark suite. Detailed settings and qualitative results are provided in the supplementary material. 
    
    Overall, MMATR-X, using a single modality-shared encoder, achieves leading transfer performance across diverse downstream tasks in both optical and SAR modalities. Notably, it is pretrained on approximately 1M optical and SAR samples, which is smaller than the pretraining datasets used by several compared models, despite the competitive performance observed across the evaluation suite.
    
    \begin{table}[tb]
        \centering      
        \caption{Performance comparison on optical tasks. We adhere to the configurations of \cite{sun2022ringmo,guo2024skysense,wang2025harnessing}. The training ratio for AID and RESISC45 classification is 20\% and 10\%,  respectively. \label{tab:optbench}}
        \vspace{-2mm}    
        \resizebox{\linewidth}{!}{
        \setlength{\tabcolsep}{3pt}
     \begin{tabular}{l|cccccc}
        \toprule
        \textbf{Model} & \makecell[c]{\textbf{AID}\\OA} & \makecell[c]{\textbf{RESISC}\\OA} & \makecell[c]{\textbf{DIOR}\\mAP$_{50}$} & \makecell[c]{\textbf{LoveDA}\\mIoU} & \makecell[c]{\textbf{DDHR}\\mIoU} & \makecell[c]{\textbf{WHU}\\mIoU}  \\ \hline
        
        \multicolumn{6}{l}{\textit{Base-level backbone}}\\ \hline
        RVSA \cite{wang2022advancing} & \ttwo{97.03} & 93.93 & 75.80 &  52.26 & 87.79 & 50.86 \\
        SelectiveMAE \cite{wang2025harnessing} & 96.78 & 93.35 & 75.70 &  \ttwo{53.05} & 87.48 & 51.48  \\    
        RingMo \cite{sun2022ringmo} & 96.90 & \ttwo{94.25} &  \ttwo{75.90} & - & - & - \\
        SatLas \cite{bastani2023satlaspretrain} &  94.96 & 92.16 &  74.10 & - & 89.74 & 51.34 \\
        GFM \cite{mendieta2023towards} & 95.47 &  92.73  &  72.84 & - & 91.55 & 51.87 \\  
        MaRS \cite{yang2026mars} & 96.84 & 92.87  & 73.70 & 52.25 & \ttwo{92.25} & \ttwo{53.18} \\
        \includegraphics[height=0.7em]{figures/LOGO.png}\textbf{MMATR-X} (ours) & \tone{97.76} & \tone{94.79}  & \tone{79.00} & \tone{55.32} & \tone{92.62} & \tone{55.58} \\ \hline        
        \multicolumn{6}{l}{\textit{Large-level backbone}} \\ \hline
        SatMAE \cite{cong2022satmae} & 95.02 & 91.72 &  70.89 & - & - & - \\
        ScaleMAE \cite{reed2023scale} & 96.44 & 92.63 & 73.81 & - & - & - \\
        SelectiveMAE \cite{wang2025harnessing} & 97.25 & 94.57 & 77.80 & 54.31 & - & - \\
        SkySense \cite{guo2024skysense} & - & 94.34 & 76.74 & - & - & - \\   
        \hline
        \multicolumn{6}{l}{\textit{Huge-level backbone}} \\ \hline
        SkySense \cite{guo2024skysense} & 97.68 & 94.85 &  78.73 & - & - & - \\
        \bottomrule
        \end{tabular}}
        \vspace{-4mm}
    \end{table}

    \subsubsection{Comparison with Optical Foundation Models} 
    We evaluate scene classification on AID \cite{AID} and RESISC45 \cite{RESISC45}, object detection on DIOR \cite{li2020object}, and semantic segmentation on LoveDA \cite{loveda}, DDHR-SK \cite{ddhr}, and WHU-OPT-SAR \cite{whuoptsar}. We follow the protocols of \cite{sun2022ringmo,guo2024skysense,wang2025harnessing} to fine-tune MMATR-X for the AID, RESISC45, DIOR, and LoveDA datasets, while DDHR-SK and WHU-OPT-SAR are evaluated by us under matched downstream settings. Detection and segmentation tasks are evaluated with Faster R-CNN \cite{ren2015faster} and UPerNet \cite{xiao2018unified} heads, respectively.

    As shown in Table~\ref{tab:optbench}, CoDe-MAE achieves the highest result among the listed Base-level models on AID, RESISC45, DIOR, and LoveDA, with 97.76 OA, 94.79 OA, 79.00 mAP$_{50}$, and 55.32 mIoU, respectively. Compared with the strongest listed Base-scale results, the differences on DIOR and LoveDA are 3.10 mAP$_{50}$ and 2.27 mIoU, respectively. For the larger models, CoDe-MAE reports higher results on DIOR and LoveDA, while SkySense-H is slightly higher on RESISC45 (94.85 vs.\ 94.79).

    \begin{table}[tb]
        \centering
        \caption{Performance comparison on SAR downstream tasks among Base-level foundation models. For 40-shot classification and detection, we follow the protocol of \cite{sarjepa,liu2025sarmae,saratrx,du2025summit}.}
        \label{tab:sarbench}
        \vspace{-2mm}
        \setlength{\tabcolsep}{3pt}
        \begin{tabular}{l|ccc}
        \toprule
        \multirow{2}{*}{\textbf{Model}} & \textbf{FUSAR-SHIP} & \textbf{MSTAR} & \textbf{SAR-ACD}  \\ 
        & OA & OA & OA  \\ \hline
        SAR-JEPA \cite{sarjepa} & 85.8 & 91.6 & 75.5  \\
        SARATR-X \cite{saratrx} & 87.7 & \ttwo{98.1} & \ttwo{76.4}  \\
        SUMMIT \cite{du2025summit} & 81.5 & 63.6 & 68.7 \\
        SARMAE \cite{liu2025sarmae} & \ttwo{89.3} & 96.7 & -  \\
        MaRS \cite{yang2026mars} & 77.7 & 75.5 & 68.4  \\
        \includegraphics[height=0.7em]{figures/LOGO.png}\textbf{MMATR-X} (ours) & \tone{89.4} & \tone{98.7} & \tone{78.3}   \\  \hline \hline

        \multirow{2}{*}{\textbf{Model}} & \textbf{SARDet-100K} & \textbf{DDHR} & \textbf{WHU} \\ 
        & mAP & mIoU & mIoU \\ \hline
        SAR-JEPA \cite{sarjepa}  & -  & 68.27 & 34.81 \\
        SARATR-X \cite{saratrx} & 57.3 & 77.54 &  42.99 \\
        SUMMIT \cite{du2025summit}  & 57.0  & 80.65 & \ttwo{44.89} \\
        SARMAE \cite{liu2025sarmae}  & \ttwo{57.9}  & 81.12 & 43.76 \\
        MaRS \cite{yang2026mars}  & 55.4 & \ttwo{83.79} & 33.92 \\
        \includegraphics[height=0.7em]{figures/LOGO.png}\textbf{MMATR-X} (ours) & \tone{61.1} & \tone{89.31} & \tone{48.64} \\
        \bottomrule
        \end{tabular}
        \vspace{-4mm}
    \end{table}
   
    \subsubsection{Comparison with SAR Foundation Models} 
    For SAR-only transfer, we evaluate 40-shot target classification on FUSAR-SHIP \cite{hou2020fusar}, MSTAR \cite{mstar}, and SAR-ACD \cite{sun2022scan}, object detection on SARDet-100K \cite{li2024sardet}, and semantic segmentation on DDHR-SK \cite{ddhr} and WHU-OPT-SAR \cite{whuoptsar}. The first four benchmarks follow the evaluation protocols used in recent SAR foundation-model studies \cite{saratrx,liu2025sarmae,du2025summit}, while DDHR-SK and WHU-OPT-SAR are reproduced and evaluated by us under matched settings.

    As shown in Table~\ref{tab:sarbench}, MMATR-X achieves 89.4, 98.7, and 78.3 OA on FUSAR-SHIP, MSTAR, and SAR-ACD, respectively. On SARDet-100K, it obtains 61.1 mAP, with differences of 3.2 points over the second-best model. For scene-level semantic segmentation, MMATR-X reaches 89.31 mIoU on DDHR-SK and 48.64 mIoU on WHU-OPT-SAR. These results cover both object-level SAR recognition and detection on target-chip or object-centric datasets, as well as scene-level dense prediction.

    \subsubsection{Comparison with Multimodal Foundation Models}

    \textbf{Linear land-cover classification.} Existing multimodal remote-sensing foundation models have largely focused on land-cover understanding and are commonly pretrained on medium-resolution Sentinel observations. Following the evaluation protocols of RS-SSL and FG-MAE \cite{swinssl,FGMAE}, we conduct 10-shot linear probing on three high-resolution datasets, PIE-RGB-SAR \cite{asanet}, DDHR-SK \cite{ddhrnet}, and WHU-OPT-SAR \cite{whuoptsar}, and three medium-resolution datasets, DFC20 \cite{dfc20}, BigEarthNet-MM \cite{BENMM}, and EuroSAT-MM \cite{FGMAE}. All six datasets provide paired optical and SAR observations, allowing the two modalities to be evaluated separately under the same protocol. For the first four datasets, the semantic segmentation annotations are converted into image-level multi-label targets and evaluated using micro mAP, while EuroSAT-MM is evaluated as single-label classification using OA. All compared models are evaluated by us under matched settings.

    As shown in Table~\ref{tab:compmm}, MMATR-X maintains competitive performance on the medium-resolution datasets despite the substantial resolution gap from its high-resolution pretraining data, indicating useful cross-resolution transfer under the frozen linear-probing setting.
    
    \textbf{Bi-temporal cross-modal change detection}. We also evaluate MMATR-X on the BRIGHT \cite{Chen2025Bright} (DFC25-T2 \cite{dfc25report} split) for building-damage assessment. This task combines pre-disaster optical imagery with post-disaster SAR observations, requiring the downstream model to jointly process cross-modal relations and temporal changes. Following the official benchmark\footnote{\url{https://github.com/ChenHongruixuan/BRIGHT}}, we use DamageFormer \cite{damageformer} as the downstream architecture and cite results of other models from \cite{yang2026mars}. As shown in Table~\ref{tab:bright}, MMATR-X achieves 44.74 mIoU, 3.25 points higher than MaRS.

    \begin{table*}[tb]
        \centering
        \caption{Performance comparison of 10-shot linear probing for land-cover classification. HR indicates that the pretraining dataset contains high-resolution imagery.\label{tab:compmm}}
        \vspace{-2mm}
        \begin{tabular}{l|c|cc|cc|cc|cc|cc|cc}
            \toprule
            \multirow{2}{*}{\textbf{Model}} &  \multirow{2}{*}{\textbf{HR}} & \multicolumn{2}{c|}{\textbf{PIE}}  & \multicolumn{2}{c|}{\textbf{DDHR-SK}} & \multicolumn{2}{c|}{\textbf{WHU}}  & \multicolumn{2}{c|}{\textbf{DFC20}}  & \multicolumn{2}{c|}{\textbf{BEN-MM}}  & \multicolumn{2}{c}{\textbf{EuroSat}}  \\   \cline{3-14} 
           &  &  Opt. & SAR &  Opt. & SAR &  Opt. & SAR &  Opt. & SAR &  Opt. & SAR &  Opt. & SAR  \\ \hline

        \multicolumn{14}{l}{\textit{Tiny-level backbone}}\\ \hline
            RS-SSL \cite{swinssl} &  & 80.12 & 79.43 & 59.06 & 76.62 & 78.47 & 79.51 & 51.56 & 54.01 & 49.17 & 49.85 & 48.00 & 40.85 \\ \hline
            
        \multicolumn{14}{l}{\textit{Small-level backbone}}\\ \hline
            DINO-MM \cite{dinomm} &  & 84.70 & 84.66 & 75.17 & 85.52 & 79.04 & \ttwo{84.08} & 47.55 & 59.61 & 52.02 & 54.54 & 50.66 & 47.39 \\ 
            
            DeCUR \cite{decur} & & 79.20 & 79.59 & 72.63 & 81.69 & 77.51 & 77.23 & 57.95 & 53.02 & 56.22 & 53.20 & 60.48 & 48.08 \\ \hline

        \multicolumn{14}{l}{\textit{Base-level backbone}}\\ \hline
            FG-MAE \cite{FGMAE} &  & 81.95 & 81.59 & 60.86 & 81.56 & 76.75 & 80.77 & 48.95 & 54.84 & 53.19 & 52.33 & 48.88 & 42.72 \\ 
            
            SatViT \cite{satvit} &  & 86.05 & 83.82 & 74.33 & 83.16 & 83.19 & 83.67 & 55.17 & 56.27 & 56.55 & 53.42 & 54.94 & 47.03 \\
            
            CROMA \cite{croma} &  & 83.03 & 81.99 & 71.69 & 84.25 & 79.78 & 82.05 & 50.39 & 55.37 & 55.90 & 54.83 & 59.12 & 47.97 \\ 
            
            DOFA \cite{xiong2024neural} & \checkmark & \ttwo{88.19} & \ttwo{86.13} & 81.57 & \ttwo{87.13} & \ttwo{85.45} & 81.44 & 62.20 & \ttwo{60.27} & 57.94 & \ttwo{56.50} & 70.25 & \ttwo{55.87} \\
            
            MaRS \cite{yang2026mars} & \checkmark & 85.93 & 79.18 & \ttwo{81.86} & 76.27 & 81.98 & 68.36 & \ttwo{65.48} & 27.82 & \ttwo{60.22} & 36.83 & \ttwo{79.79} & 32.66 \\
            
            \includegraphics[height=0.7em]{figures/LOGO.png}\textbf{MMATR-X} (Ours) & \checkmark & \tone{91.27} & \tone{88.78} & \tone{84.86} & \tone{89.20} & \tone{86.86} & \tone{87.29} & \tone{70.43} & \tone{62.55} & \tone{65.38} & \tone{57.97} & \tone{81.18} & \tone{59.88} \\             
            \bottomrule
        \end{tabular}
        \vspace{-4mm}
    \end{table*}
    
    \begin{table}[tb]
        \centering
        \caption{Performance comparison on BRIGHT (DFC25-T2) dataset. We leverage the Siamese-style DamageFormer \cite{damageformer} framework.}
        \label{tab:bright}
        \vspace{-2mm}    
        \begin{tabular}{l|c}
            \toprule
            \multirow{2}{*}{\textbf{Model}} & \textbf{DFC25-T2}  \\ 
            &  mIoU  \\ \hline    
            DOFA \cite{xiong2024neural} & 37.59 \\
            SatMAE \cite{cong2022satmae} & 29.39 \\
            GFM \cite{mendieta2023towards} & 28.24 \\
            SatLas \cite{bastani2023satlaspretrain} & 37.06 \\
            MaRS \cite{yang2026mars} & \ttwo{41.49} \\
            \includegraphics[height=0.7em]{figures/LOGO.png}\textbf{MMATR-X} (Ours) & \tone{44.74} \\ 
            \bottomrule
        \end{tabular}
        \vspace{-4mm}
    \end{table}
    
    \subsubsection{Data-overlap Audit}\label{sec:dataoverlapautid}
    
    We audit OSPretrain-1M for potential leakage into downstream evaluation sets. In remote-sensing pretraining, overlap in sensor family, geographic region, spatial resolution, or acquisition period does not by itself constitute benchmark contamination; the more direct concern is whether evaluation samples, annotations, or duplicate/near-duplicate observations are exposed during pretraining. OSPretrain-1M uses no downstream annotations, and identifiable official evaluation partitions are excluded whenever explicit split information is available. We further screen the downstream test sets against OSPretrain-1M using cryptographic and perceptual hashes. Publicly available provenance information, including sensor, resolution, and geographic coverage, is summarized in the supplementary material; however, finer-grained geographic or temporal overlap cannot always be verified because many benchmarks do not provide image-level coordinates, timestamps, or scene identifiers.

    No exact or near-duplicate overlap is detected for most benchmarks. One exception occurs between the unpaired SARDet-180K \cite{saratrx} subset used in pretraining and the SARDet-100K \cite{li2024sardet} test set, where 160 overlapping images are identified. These overlaps originate from SAR-AIRCraft-1.0 \cite{zhirui2023sar}, a source dataset shared by both collections. Auditing its original release reveals duplicate images crossing the original training and test partitions; this inconsistency therefore propagates to both derived datasets despite their preservation of the source splits. After removing the identified overlapping images from the SARDet-100K test set, mAP decreases by only 0.1 points. We report this decontaminated result together with the standard full-test result, which is retained for direct comparison with prior work.
    
    \subsection{Controlled Experiments}

    We conduct a set of controlled experiments, which examine three aspects: whether the proposed framework remains effective when representative optical-SAR joint pretraining methods are trained with the same backbone and data, the computational overhead introduced by its additional objectives, and whether the same design can be applied to other heterogeneous modality pairs.
    
    \subsubsection{Matched Optical-SAR Pretraining}\label{sec:matched_optsar}
    
    We re-implement MAE \cite{mae}, RS-SSL \cite{swinssl}, SatViT \cite{satvit}, DeCUR \cite{decur}, msGFM \cite{msgfm}, and MaRS \cite{yang2026mars} using the same iTPN backbone. Since several comparison methods do not naturally support unpaired observations, all methods are pretrained exclusively on the paired subset of OSPretrain-1M, with the same number of epochs and training protocol. This ensures that differences in performance are not confounded by access to additional unpaired data. Since CoDe-MAE additionally employs DINO-v3 Sat for knowledge distillation, we further incorporate the same teacher and distillation objective into each baseline to control for knowledge distillation. The resulting DeCUR variant did not converge because of gradient explosion and is therefore omitted from the distillation comparison. 
    
    We evaluate the resulting representations on DDHR-SK and WHU-OPT-SAR using optical-only, SAR-only, and joint optical-SAR semantic segmentation. For WHU-OPT-SAR, 20\% of the training labels are used. All downstream settings are kept identical across methods. For multimodal segmentation, optical and SAR features are concatenated along the channel dimension before the segmentation head, without an additional learnable fusion module.

    \begin{table}[tbp]
    \centering
    \caption{Controlled comparison across optical-SAR joint pretraining methods under the same pretraining datasets, backbone, and downstream protocol. ${^{\dag}}$Separate encoders for optical and SAR inputs. ${^{\ddag}}$The MAE baseline is pretrained separately on single-modality data. The DeCUR variant with DINO-v3 Sat distillation is not reported because training diverged due to gradient explosion under this matched configuration.}
    \label{tab:controlled_optsar}
    \vspace{-2mm}
    \resizebox{\linewidth}{!}{
    \begin{tabular}{l|ccc|ccc}
    \toprule
    \multirow{2}{*}{\textbf{Method}} & \multicolumn{3}{c|}{\textbf{DDHR-SK}} & \multicolumn{3}{c}{\textbf{WHU-OPT-SAR}} \\
    & Opt. & SAR & Both & Opt. & SAR & Both \\
    \hline
    Random init. & 79.36 & 72.97 & 84.40 & 40.62  & 36.08 & 44.78 \\ 
    MAE$^{{\dag}}$$^{{\ddag}}$ \cite{mae} & 88.74 & 83.19 & 90.68 & 53.02 & 45.93 & 53.56 \\ 
    \hline
    RS-SSL$^{{\dag}}$ \cite{swinssl} & 85.54 & 81.04 & 88.38 & 48.67 & 43.82 & 50.41 \\
    SatViT \cite{satvit} & 86.94 & 80.99 & 89.52 & 51.31 & 44.51 & 52.43  \\
    DeCUR$^{{\dag}}$ \cite{decur} & 89.93 & 85.59 & 92.08 & 50.91 & 45.40 & 52.19 \\
    msGFM \cite{msgfm} & 86.60 & 80.67 & 88.98 & 50.42 & 44.25 & 52.11 \\
    MaRS$^{{\dag}}$ \cite{yang2026mars} & 82.80 & 66.35 & 84.93 & 46.15 & 32.58 & 47.82 \\ \hline
    \multicolumn{7}{l}{\textit{with DINO-v3 Sat (ViT-L) distillation}} \\ \hline
    RS-SSL$^{{\dag}}$ \cite{swinssl} & \tone{91.01} & 82.44 & 92.18 &  53.30&  44.80 & 53.62 \\
    SatViT \cite{satvit} & 90.69 & 84.10 & 91.85 & \ttwo{53.74} & 41.65 & \ttwo{54.06} \\
    msGFM \cite{msgfm} & \ttwo{90.86} & \ttwo{86.09} & \ttwo{92.70} & 53.40 & \ttwo{45.94} & 54.00 \\
    MaRS$^{{\dag}}$ \cite{yang2026mars} & 88.31 & 69.74 & 89.17 & 52.40 & 32.38 & 52.86 \\
    \textbf{CoDe-MAE} (Ours) & {90.70} & \tone{87.03} & \tone{92.93} & \tone{53.83} & \tone{46.88} & \tone{54.23} \\ 
     \bottomrule
    \end{tabular}}
    \vspace{-4mm}
    \end{table}    

    As shown in Table~\ref{tab:controlled_optsar}, CoDe-MAE achieves the best performance in five of the six evaluated settings. Its advantage is particularly consistent for SAR-only and joint-modal transfer, where it obtains the best results on both datasets. CoDe-MAE shares an encoder between the optical and SAR branches. Although DINOv3 supervision is applied only to the optical branch, its gradients update the shared parameters used to represent both modalities. SatViT and msGFM provide comparable pathways under the matched distillation setting; however, CoDe-MAE achieves the strongest SAR-only and joint-modal performance on both DDHR-SK and WHU-OPT-SAR. This comparison suggests that, beyond parameter sharing, the proposed cross-modal objectives help translate optical teacher supervision into effective SAR representations.
    
    \subsubsection{Computational Overhead}

\begin{figure}[tb]
        \centering
        \includegraphics[width=\linewidth]{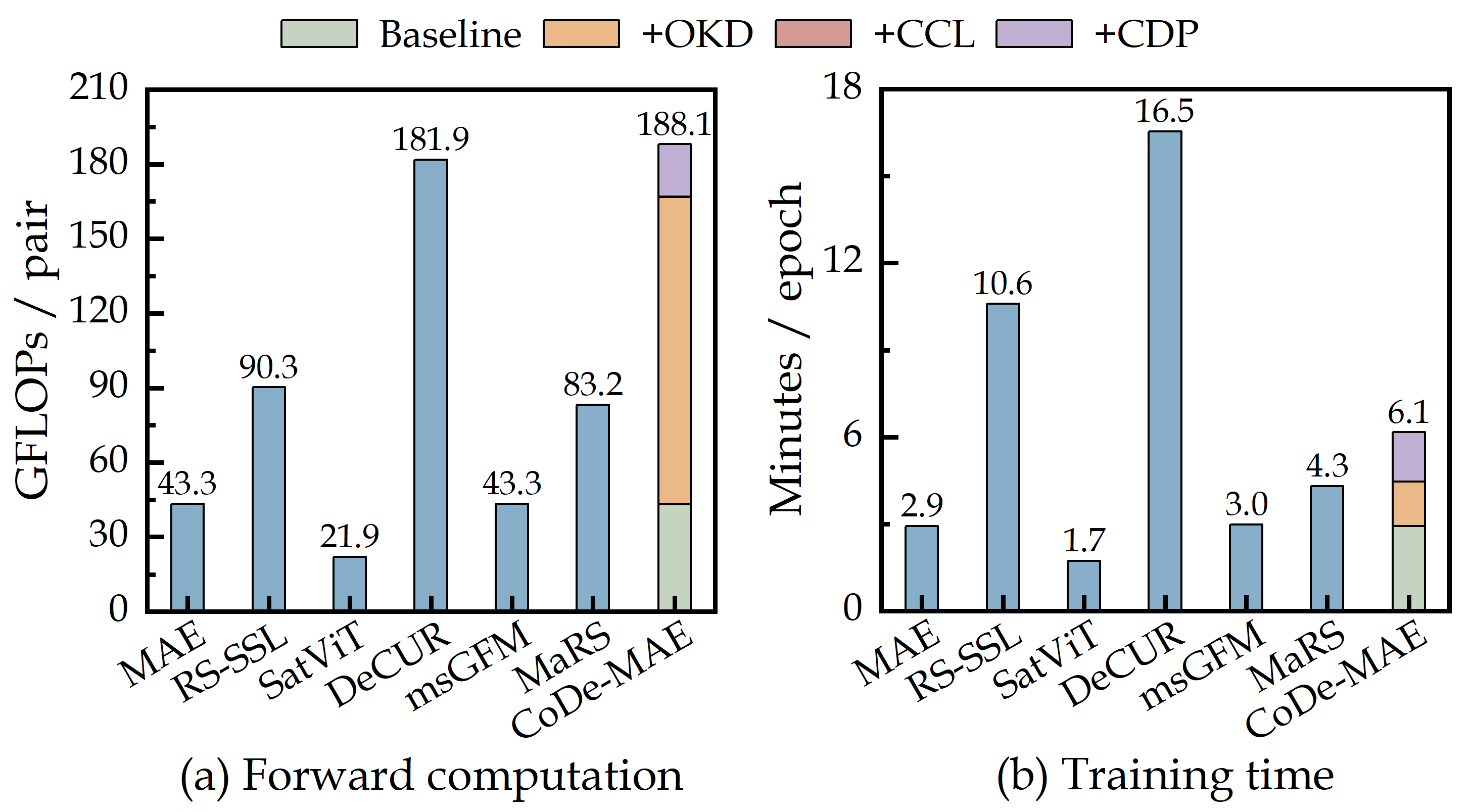}
        \vspace{-6mm}
        \caption{Computational cost of the controlled optical-SAR pretraining methods. (a) Forward computation per paired sample, measured in GFLOPs. (b) Mean wall-clock training time per epoch on eight RTX 4090D GPUs. For practical memory utilization, each method uses the largest per-GPU batch size supported by the available memory: MAE, SatViT, msGFM, MaRS, and CoDe-MAE use 128 pairs per GPU, whereas RS-SSL and DeCUR use 32 pairs per GPU. The cost of CoDe-MAE is decomposed into its base masked-autoencoding path and the additional computation introduced by OKD, CCL, and CDP.}
    \label{fig:efficiency}
        \vspace{-4mm}
    \end{figure}

    We further compare the computational complexity and practical training cost of the controlled pretraining methods. Fig.~\ref{fig:efficiency}(a) reports the forward computation per paired sample in GFLOPs, providing a batch-size-independent measure of model computation. Fig.~\ref{fig:efficiency}(b) reports the measured wall-clock training time per epoch on the same eight RTX 4090D GPUs. For the wall-clock comparison, each method uses the largest per-GPU batch size that fits within the available memory. Specifically, MAE, SatViT, msGFM, MaRS, and CoDe-MAE use 128 pairs per GPU because their encoders process only the visible tokens of masked inputs, whereas RS-SSL and DeCUR process the complete token sequence and use 32 pairs per GPU. This protocol fixes the hardware budget while allowing each method to use its practical memory-efficient configuration. The resulting epoch time therefore reflects end-to-end training cost under the available hardware rather than per-iteration latency at an artificially matched batch size.

    For CoDe-MAE, we further decompose the computation into the base masked-autoencoding path and the additional components introduced by OKD, CCL, and CDP. The largest increase in forward GFLOPs comes from the frozen DINO-v3 teacher used by OKD. Because the teacher requires forward inference but no gradient computation, this increase in forward computation does not translate proportionally into training time. In the measured wall-clock cost, CDP introduces a more noticeable overhead due to its additional reconstruction decoder, whereas the incremental cost of CCL is comparatively small. This decomposition distinguishes the cost of teacher supervision from that of the two cross-modal objectives. All additional CoDe-MAE modules are discarded after pretraining, so the controlled methods have identical downstream inference cost; the overhead reported here is therefore confined to pretraining.

    The additional cost could potentially be reduced through objective scheduling. For example, msGFM \cite{msgfm} probabilistically alternates self-reconstruction and cross-modal prediction rather than evaluating both objectives at every update. Incorporating a similar selective optimization strategy into CoDe-MAE provides a possible direction for improving training efficiency.
   
    \subsubsection{Generality to other Heterogeneous Modality Pairs}\label{demndepthcontroll}

\begin{figure}[tbp]
        \centering
        \includegraphics[width=\linewidth]{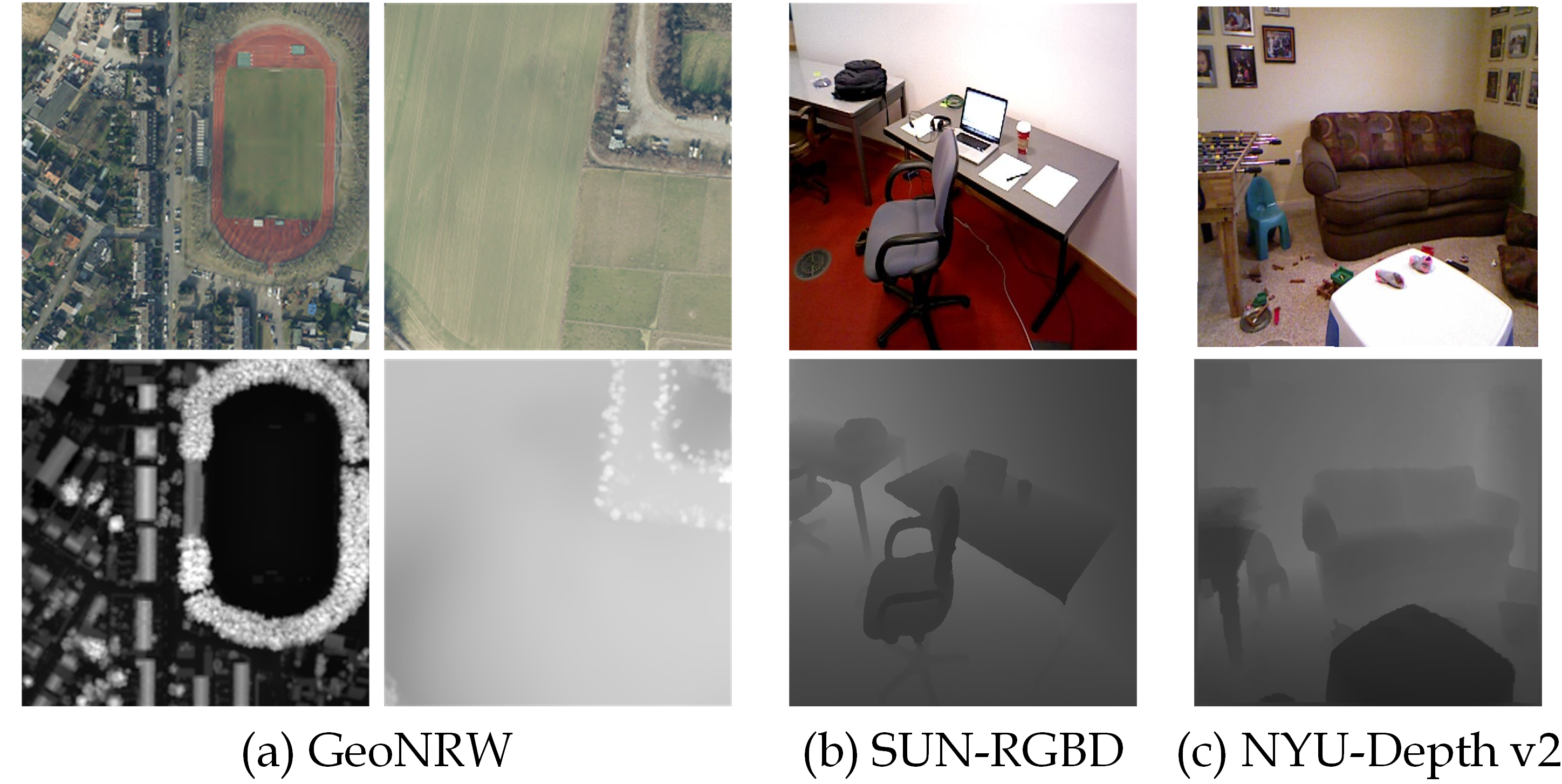}
        \vspace{-6mm}
        \caption{Representative paired samples used in the additional cross-modality experiments. (a) Aerial optical imagery and corresponding DEM data from GeoNRW \cite{geonrw}. (b) RGB images and depth maps from SUN RGB-D \cite{song2015sun}, used for RGB-depth pretraining. (c) RGB images and depth maps from NYU-Depth v2 \cite{nyudv2}, used for downstream transfer evaluation.}
    \vspace{-2mm}
    \label{fig:generalization_datasets}
    \end{figure}

    Following \cite{decur}, we conduct two additional controlled experiments involving different modality pairs. The first setting considers aerial optical imagery and DEMs from GeoNRW \cite{geonrw}. The DEMs are generated from high-resolution airborne LiDAR observations and provide geo-referenced surface-elevation information complementary to optical appearance. We construct 116,768 paired optical-DEM patches for 100-epoch pretraining and use 1\% of the labeled training data for downstream fine-tuning. The second setting considers RGB-depth pretraining on SUN RGB-D \cite{song2015sun}, where depth maps encode camera-relative distances to visible surfaces. We pretrain on 10,335 RGB-depth pairs for 200 epochs and transfer the models to NYU-Depth v2 \cite{nyudv2}. Examples are shown in Fig.~\ref{fig:generalization_datasets}. DEM and depth primarily encode geometric/structural information rather than conventional appearance cues, and in these settings they serve mainly as complementary modalities to optical/RGB imagery. We therefore evaluate both the primary modality alone and the corresponding multimodal input.
    
    Table~\ref{tab:crossmodal_generalization} reports the results with and without DINO-v3-based distillation. Under the matched distillation setting, CoDe-MAE achieves the best multimodal performance on both modality pairs, while also ranking first or second for primary-modality transfer. In particular, it obtains 55.88 with optical+DEM on GeoNRW and 63.12 with RGB+depth on NYU-Depth v2, while maintaining competitive optical-only and RGB-only performance. These results provide additional evidence that the pair-conditioned alignment and selective reconstruction principles extend beyond optical-SAR to other heterogeneous modality combinations.

    \begin{table}[tbp]
        \centering
        \caption{Controlled evaluation on two additional heterogeneous modality pairs. Optical-DEM models are pretrained on GeoNRW for 100 epochs and evaluated using 1\% of the labeled training data. RGB-depth models are pretrained on SUN RGB-D for 200 epochs and transferred to NYU-Depth v2. RGB-depth CDP uses Canny-edge targets. Results are reported for the primary modality alone and for multimodal input. ${^{\dag}}$Separate encoders for optical and SAR inputs. ${^{\ddag}}$The MAE baseline is pretrained separately on single-modality data.}
        \label{tab:crossmodal_generalization}
        \vspace{-2mm}
        \begin{tabular}{l|cc|cc}
        \toprule
        \multirow{2}{*}{\textbf{Method}} & \multicolumn{2}{c|}{\textbf{GeoNRW}} & \multicolumn{2}{c}{\textbf{NYU-Depth v2}} \\
        & Opt. & Both  & RGB & Both\\
        \hline
        Random init. & 23.03 & 30.36 & 11.28 & 19.26 \\
        MAE$^{{\dag}}$$^{{\ddag}}$ \cite{mae} & 50.98  &  52.95 & 42.32 & 45.93  \\ \hline
        RS-SSL$^{{\dag}}$ \cite{swinssl} & 43.70 &  45.65 & 48.74 & 49.17 \\
        SatViT \cite{satvit} & 46.92 & 49.32 & 26.20 & 35.29 \\
        DeCUR$^{{\dag}}$ \cite{decur} & 52.63 & 54.52 & 47.12 & 49.08 \\
        msGFM \cite{msgfm} & 50.09 & 52.21 & 29.96 & 35.80  \\
        MaRS$^{{\dag}}$ \cite{yang2026mars} & 43.84 & 46.39 & 27.55 & 34.60 \\ \hline 
        \multicolumn{5}{l}{\textit{with DINO-v3 Sat/Lvd (ViT-L) distillation}} \\ \hline
        RS-SSL$^{{\dag}}$ \cite{swinssl} & 52.24 & 52.78 & 60.77 &  \ttwo{61.17} \\
        SatViT \cite{satvit} & 52.75 & 53.67 & 57.95 & 58.40 \\
        DeCUR$^{{\dag}}$ \cite{decur} & 53.13 & \ttwo{55.25} & 47.08 & 49.04 \\
        msGFM \cite{msgfm} & \tone{54.41} & 54.89 & \ttwo{60.91} & \ttwo{61.17} \\
        MaRS$^{{\dag}}$ \cite{yang2026mars} & 51.49 & 51.62 & 49.29 & 51.01 \\
        \textbf{CoDe-MAE} (Ours) & \ttwo{54.40} & \tone{55.88} & \tone{63.05} & \tone{63.12} \\
        \bottomrule
        \end{tabular}
    \vspace{-4mm}
    \end{table}
    
    \subsection{Ablation Study} \label{sec:abalation}

    We conduct controlled ablations across optical-SAR, optical-DEM, and RGB-depth pairs to examine the contributions of the proposed components, the design of cross-modal alignment and prediction, and the use of paired and unpaired pretraining data of OSPretrain-1M. Note that the optical-SAR ablations use the full OSPretrain-1M dataset, including both its paired and unpaired subsets, except for the ablation on dataset composition.

    \subsubsection{Component Ablation}

    Table~\ref{tab:ablationcclcdp} evaluates the contributions of OKD, CCL, and CDP within the same shared-backbone architecture. OKD provides a stable improvement over the base masked-autoencoding model across all three modality pairs, establishing a strong teacher-regularized baseline. 
    
    For the configurations reported in Table~\ref{tab:ablationcclcdp}, adding either CCL or CDP improves the measured performance over OKD, and their combination gives the highest values in every evaluated setting. As can be seen in Section~\ref{demndepthcontroll}, additional cross-modal supervision does not always improve upon this baseline: most existing joint-pretraining methods yield lower primary-modality transfer than OKD alone. The observed gains provide empirical support for combining pair-conditioned alignment and selective cross-modal prediction with optical teacher supervision.
    
    \begin{table}[tbp]
        \centering
        \caption{Component ablation of OKD, CCL, and CDP across optical-SAR, optical-DEM, and RGB-depth pretraining. For the additional modality pairs, we report transfer performance using the primary visual modality. The MAE baseline is pretrained separately on single-modality data.}        
        \label{tab:ablationcclcdp}
        \vspace{-2mm}    
        \begin{tabular}{ccc|cc|c|c}
            \toprule
            \multirow{2}{*}{\textbf{OKD}} & \multirow{2}{*}{\textbf{CCL}} & \multirow{2}{*}{\textbf{CDP}} & \multicolumn{2}{c|}{\textbf{DDHR-SK}} & \textbf{GeoNRW} & \textbf{NYU-D} \\
            &  &   & Opt. & SAR & Opt. & RGB \\ 
            \hline
            \multicolumn{3}{l|}{\,\,\textit{MAE baseline}} & 88.61 & 83.01 & 50.98 & 42.32 \\ \hline
            & & & 88.77 & 84.40 & 50.67 & 35.03 \\ 
            \checkmark &  &  & 91.25 & 86.77 & 53.67 & 62.44 \\
            \checkmark &  \checkmark &  & {91.40} & 86.85 & 53.78 & \ttwo{62.84} \\
            \checkmark &  & \checkmark & \ttwo{91.42} & \ttwo{87.27} & \ttwo{54.33} & 62.77 \\
           \checkmark &  \checkmark & \checkmark & \tone{91.56} &\tone{87.38} & \tone{54.40} & \tone{63.05} \\ 
            \bottomrule
        \end{tabular}
    \vspace{-2mm}
    \end{table}

    \begin{table}[tp]
        \centering
        \caption{Ablation of the representation used for contrastive alignment. Values denote performance changes relative to the OKD baseline. We compare direct alignment of encoder features, counterpart-independent linear and nonlinear projection heads, and the proposed pair-conditioned alignment through CCL.}
        \label{tab:ablationconttarget}
        \vspace{-2mm}    
        \begin{tabular}{l|cc|c|c}
            \toprule
            \multirow{2}{*}{\makecell[l]{\textbf{Alignment}\\\textbf{representation}}} & \multicolumn{2}{c|}{\textbf{DDHR-SK}} & \textbf{GeoNRW} & \textbf{NYU-D}\\
             & Opt. & SAR & Opt. & RGB \\ \hline
            \cellcolor{blue!10}\textit{OKD baseline} & \cellcolor{blue!10}91.25 & \cellcolor{blue!10}86.77 & \cellcolor{blue!10}53.67 & \cellcolor{blue!10}62.44 \\ \hline
             Raw embedding  & \cellcolor{green!10}-0.31 & \cellcolor{green!10}-0.40 & \cellcolor{green!10}-0.46 & \cellcolor{green!10}-0.23 \\ 
            Linear projection &  \cellcolor{green!10}-0.22 &  \cellcolor{green!10}-0.59 & \cellcolor{green!10}-0.32  & \cellcolor{red!10}+0.02  \\
             MLP projection & +0.00 &  \cellcolor{green!10}-0.37 & \cellcolor{red!10}\ttwo{+0.06} & \cellcolor{red!10}\ttwo{+0.17} \\
             CCL  & \cellcolor{red!10}\tone{+0.15} & \cellcolor{red!10}\tone{+0.08} & \cellcolor{red!10}\tone{+0.11} & \cellcolor{red!10}\tone{+0.40}\\    
            \bottomrule
        \end{tabular}
    \vspace{-4mm}
    \end{table}

    \subsubsection{Contrastive Alignment Space} 

    Table~\ref{tab:ablationconttarget} compares different representations on which the contrastive objective is applied. Directly aligning the encoder representations generally degrades transfer performance, and introducing linear or nonlinear projection heads does not produce consistent improvements across the three modality pairs. Increasing the capacity of the counterpart-independent projector alleviates the degradation in some settings, but the resulting gains remain modality-dependent.

    CCL is the only tested alignment design with positive measured gains over the OKD baseline in every reported setting, including optical and SAR transfer on DDHR-SK and the primary-modality transfer on GeoNRW and NYU-Depth v2. These results provide empirical support for pair conditioning in the evaluated settings and are consistent with the motivation in Section~\ref{sec:analysis_alignment} to make paired information available to the alignment function.

    \begin{table}[tbp]
        \centering
        \caption{Ablation of the cross-modal prediction target. Values denote performance changes relative to the OKD baseline. Targets are grouped by spatial smoothing, spectral projection, combined transformation, and structural abstraction. $^{\dagger}$The optical-SAR Canny configuration diverged during pretraining with NaN loss, and therefore has no downstream result.}
        \label{tab:ablationcrtarget}
        \vspace{-2mm}    
        \begin{tabular}{l|cc|c|c}
            \toprule
            \multirow{2}{*}{\makecell[l]{\textbf{Cross-modal}\\\textbf{prediction target}}} & \multicolumn{2}{c|}{\textbf{DDHR-SK}} & \textbf{GeoNRW} & \textbf{NYU-D} \\
             & Opt. & SAR & Opt. & RGB \\ \hline
            \cellcolor{blue!10}\textit{OKD baseline} & \cellcolor{blue!10}91.25 & \cellcolor{blue!10}86.77 & \cellcolor{blue!10}53.67 & \cellcolor{blue!10}62.44 \\ \hline
             Raw pixels & \cellcolor{green!10}-0.20 & \cellcolor{red!10}+0.13 & \cellcolor{red!10}+0.24& \cellcolor{green!10}-2.38 \\ \hline
             
            \multicolumn{5}{l}{\textit{Spatial smoothing}} \\ \hline
             Median filtering  & \cellcolor{green!10}-0.28 & \cellcolor{green!10}-0.01 & \cellcolor{red!10}+0.34 & \cellcolor{green!10}-2.92 \\
             Gaussian filtering & +0.00 & \cellcolor{green!10}-0.11 & \cellcolor{red!10}+0.03& \cellcolor{green!10}-3.31 \\ \hline
             
        \multicolumn{5}{l}{\textit{Spectral projection}} \\ \hline
           Grayscale (luminance)  & \cellcolor{red!10}\tone{+0.17} & \cellcolor{red!10}\tone{+0.50} & \cellcolor{red!10}\tone{+0.66} &  \cellcolor{green!10}-0.92 \\ \hline
           
         \multicolumn{5}{l}{\textit{Combined degradation}} \\     \hline
          Gaussian + Grayscale  & \cellcolor{green!10}{-0.07} & \cellcolor{green!10}{-0.05} & \cellcolor{red!10}{+0.54}& \cellcolor{green!10}-1.86 \\ \hline
          
        \multicolumn{5}{l}{\textit{Structural abstraction}} \\ \hline
           Canny & \multicolumn{2}{c|}{Diverged$^{\dagger}$} & \cellcolor{green!10}{-0.11} & \cellcolor{red!10}\tone{+0.33}  \\
            HOG  & \cellcolor{red!10}\ttwo{+0.15} & \cellcolor{red!10}+0.17 & \cellcolor{red!10}\tone{+0.66}& \cellcolor{green!10}-0.76 \\
           MGF & \cellcolor{green!10}-0.01 & \cellcolor{red!10}\ttwo{+0.30} & \cellcolor{red!10}\ttwo{+0.57} & \cellcolor{green!10}-1.63 \\
            \bottomrule
        \end{tabular}
    \vspace{-4mm}
    \end{table}
    
    \subsubsection{Cross-modal Prediction Target}\label{sec:ablationcdp}

    Table~\ref{tab:ablationcrtarget} examines how the choice of cross-modal prediction target affects pretraining. We first consider optical-SAR, our primary setting. Predicting the raw counterpart or applying generic spatial smoothing provides little or negative improvement over the OKD baseline. In contrast, the simple luminance-based grayscale target improves both optical and SAR transfer and performs better than the evaluated handcrafted structural targets such as HOG \cite{FGMAE} and MGF \cite{saratrx}; the Canny configuration diverges during optical-SAR pretraining and therefore has no downstream result.

    The preferred target nevertheless varies across modality pairs. For optical-DEM, most evaluated targets improve upon the OKD baseline, with grayscale and HOG giving the strongest gains. This broader tolerance to target transformation may reflect the stronger structural correspondence between optical imagery and DEM, which can reduce cross-modal ambiguity relative to optical-SAR. RGB-depth exhibits a different pattern: most reconstruction targets still degrade transfer, but the luminance-based grayscale target is substantially less harmful than spatial smoothing alone. Moreover, combining grayscale with Gaussian smoothing partially mitigates the degradation caused by smoothing, while Canny edges provide the only positive gain among the evaluated targets. 
        
    Overall, these results indicate that effective target transformation depends on the form of cross-modal correspondence rather than simply on how much information is removed, consistent with the predictability-retention principle in Section~\ref{sec:analysiscdp}. Adaptive degradation is a promising extension, but learning it solely from reconstruction loss may favor easily predictable yet uninformative targets, as lower pretraining loss does not necessarily translate to better downstream transfer \cite{feng2026tessera}. In the recent SARATR-X v2 \cite{saratrxv2}, a SAR-specific prior makes such adaptation feasible by restricting the target space to multi-scale gradient features and learning only their combination. For general multimodal pretraining, future extensions may consider adaptive selection or combination of target transformations conditioned on correspondence, learnable degradation functions, and surrogate objectives \cite{liu2026s4st} that better reflect representation utility than reconstruction fidelity alone.
    
    \subsubsection{Paired and Unpaired Pretraining Data}

    \begin{table}[tbp]
        \centering
        \caption{Effect of incorporating unpaired observations from OSPretrain-1M. Both settings use the complete CoDe-MAE objective and differ only in the pretraining corpus: {Paired-only} uses the registered optical-SAR subset, whereas {Full set} additionally includes unpaired optical and SAR samples with broader modality-specific and target-centric coverage. We report mIoU for DDHR-SK segmentation, OA for RESISC45 classification, mAP$_{50}$ and mAP for DIOR and SARDet-100K detection, respectively. SARDet-100K is fine-tuned using the 1$\times$ schedule (12 epochs) for efficient ablation, rather than the 36-epoch schedule used in the MMATR-X comparison with existing foundation models.}
        \label{tab:ablationdata}
        \vspace{-2mm}    
        \begin{tabular}{l|cc|c|c|c}
            \toprule
            \multirow{2}{*}{\textbf{OSPretrain-1M}} & \multicolumn{2}{c|}{\textbf{DDHR-SK}} & \textbf{RESISC} & \textbf{DIOR} & \textbf{SARDet} \\
             & Opt. & SAR & Opt. & Opt. & SAR \\ \hline
             Paired-only & 90.70 & 87.03 & 93.99 & 76.30 & 52.40 \\
             Full set & {91.56} & 87.38 & 94.54 & 77.80 & 53.50 \\
            \bottomrule
        \end{tabular}
    \vspace{-4mm}
    \end{table}

    We finally examine the contribution of the unpaired observations in OSPretrain-1M. Recall that the paired subset mainly provides registered scene-level optical-SAR observations and supports the complete CoDe-MAE objective, whereas the unpaired subset broadens the modality-specific and target-centric content, including object-focused imagery for which reliable cross-modal pairing is difficult to establish. These samples participate only in objectives that do not require cross-modal correspondence, \ie masked reconstruction and, for optical inputs, OKD. We compare the complete CoDe-MAE model pretrained on the paired subset alone with the same model pretrained on the full OSPretrain-1M dataset.

    As shown in Table~\ref{tab:ablationdata}, incorporating the unpaired data consistently improves transfer over paired-only pretraining. The flexible objective design of CoDe-MAE allows paired samples to provide explicit cross-modal supervision while still exploiting unpaired observations through modality-specific objectives. The gains are observed across optical and SAR classification/detection tasks and further extend to dense semantic understanding on DDHR-SK. These results show that broader modality-specific and target-centric data can effectively complement paired cross-modal supervision.
    
    \subsection{Empirical Analysis of Cross-modal Behavior}\label{sec:mechanalysis}

    Beyond downstream transfer performance, we further examine several empirical properties of models trained with CCL and CDP. We consider representation geometry, cross-modal prediction behavior, and base-representation similarity across different levels of image-level optical-SAR correspondence. These analyses provide complementary empirical evidence for understanding the behavior of the two cross-modal objectives.
    
    \subsubsection{Representation Geometry} 

\begin{figure}[tbp]
        \centering
        \includegraphics[width=\linewidth]{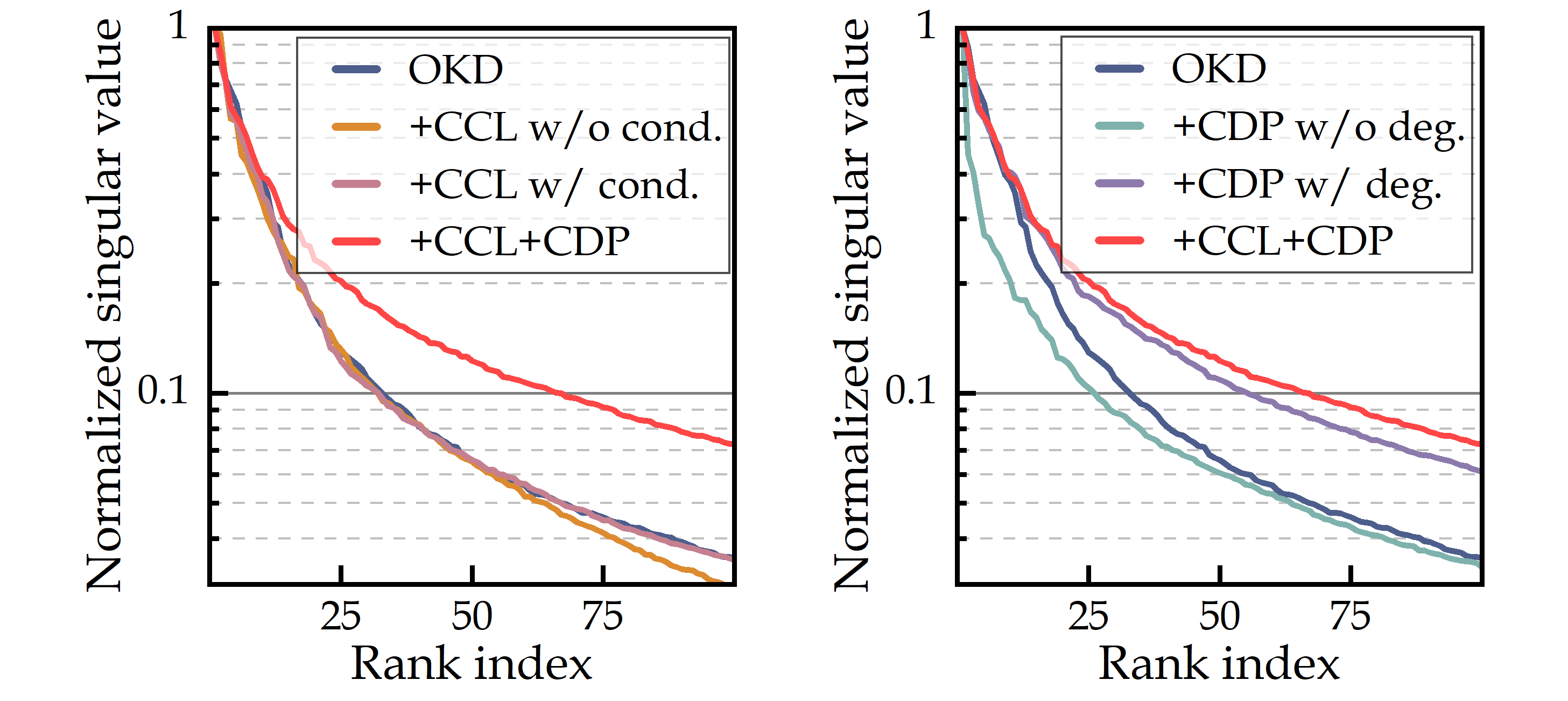}
        \vspace{-6mm}
        \caption{Normalized singular-value spectra of SAR representations. The left panel compares contrastive alignment variants and the right panel compares cross-modal prediction variants. Pair conditioning and target degradation reduce the spectral concentration observed with their direct counterparts, while the full CoDe-MAE model retains the broadest spectrum.}
        \label{fig:spectral}
        \vspace{-2mm}
    \end{figure}

    We first examine how the cross-modal objectives affect the geometry of the learned representations. Fig.~\ref{fig:spectral} shows the normalized singular-value spectra of SAR features, which are used here because the SAR branch does not receive direct teacher supervision. Direct contrastive alignment and raw cross-modal prediction are associated with faster spectral decay, indicating that feature variation becomes more concentrated in the leading directions. Pair conditioning and target degradation both alleviate this tendency, while the full CCL+CDP model retains the broadest spectrum. This indicates that CCL and CDP mitigate spectral concentration, yielding SAR representations with feature variation distributed more broadly across singular directions.
    
\begin{figure}[tbp]
        \centering
        \includegraphics[width=\linewidth]{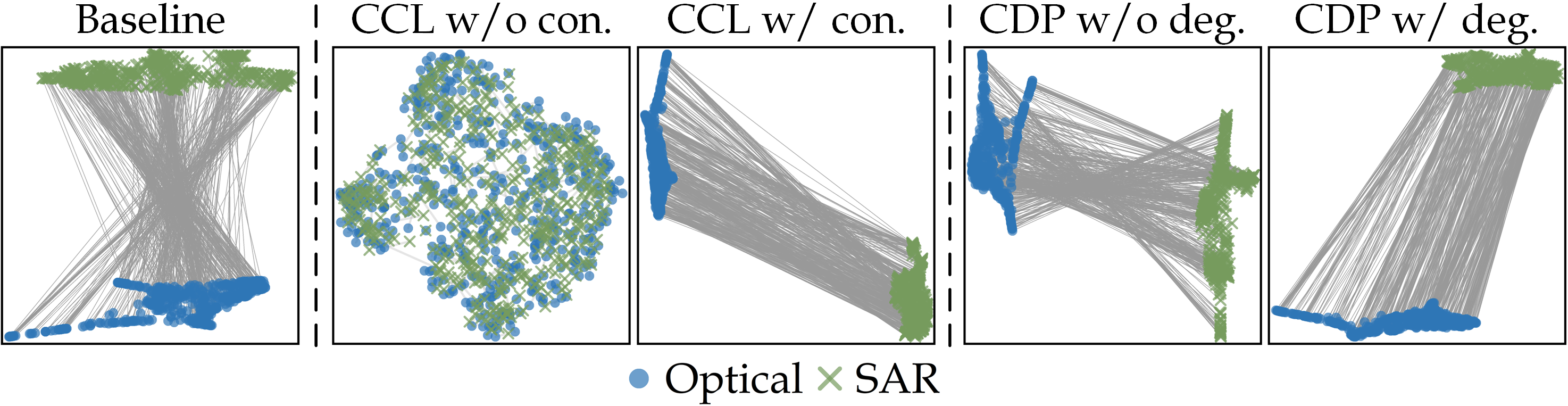}
        \vspace{-6mm}
        \caption{UMAP visualization of paired optical and SAR representations, with gray lines connecting corresponding samples. CCL and CDP show distinguishable modality distributions in the projected space while producing more consistently organized paired connections than their direct counterparts, providing a qualitative view of cross-modal representation geometry.}
        \label{fig:analysis_umap}
        \vspace{-4mm}
    \end{figure}

    Fig.~\ref{fig:analysis_umap} further visualizes the cross-modal representation geometry using UMAP \cite{umap}, with lines connecting paired optical and SAR samples. The OKD baseline shows clear modality separation but highly intersecting paired connections. Direct contrastive alignment largely mixes the two modality distributions, whereas raw-target cross-modal prediction produces more fragmented projected structures. In contrast, CCL and CDP show distinguishable modality distributions in the projected space while exhibiting more consistently organized paired connections. The modality separation suggests that optical and SAR representations remain distinguishable, while the organization of paired connections provides a qualitative view of their cross-modal geometric correspondence.

    \subsubsection{Effect of Target Degradation on Cross-modal Prediction}

\begin{figure}[tbp]
        \centering
        \includegraphics[width=\linewidth]{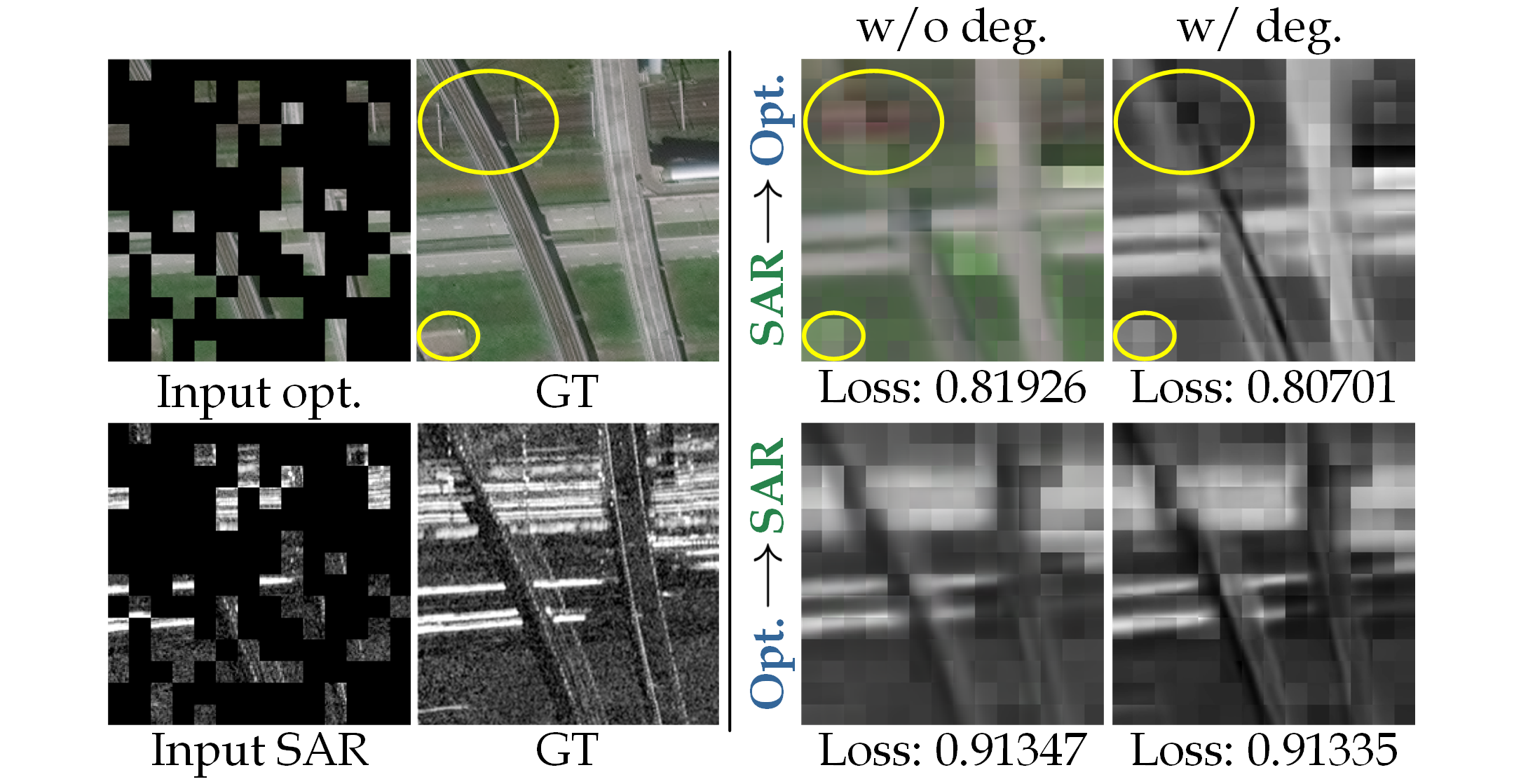}
        \vspace{-6mm}
        \caption{Qualitative comparison of cross-modal prediction with and without target degradation. In SAR-to-optical prediction, the full RGB target leads to a prediction affected by surrounding appearance and an incorrectly recovered bridge region, whereas the luminance-based grayscale target recovers the bridge structure more faithfully. With the shared encoder, optical-to-SAR prediction also shows clearer structural contrast after applying CDP.}
        \label{fig:analysis_recon}
        \vspace{-2mm}
    \end{figure}

    Fig.~\ref{fig:analysis_recon} illustrates how target degradation changes cross-modal prediction behavior. In SAR-to-optical prediction, when the full RGB target is used, the masked bridge region is predicted with a brown appearance similar to the surrounding ground, although the bridge is gray in the ground truth. The prediction therefore follows the local visual context rather than the actual bridge appearance, and its structure is also poorly recovered. With the luminance-based grayscale target, the bridge structure is predicted more faithfully. This example is consistent with the motivation of reducing chromatic prediction requirements while retaining luminance-based structural cues.

    Because the optical and SAR branches share the encoder, changing the prediction target in one direction also changes the supervision received by parameters used for both modalities. Consistent with this coupling, optical-to-SAR prediction with CDP exhibits higher contrast and clearer structural details than its raw-target counterpart. Together with the target ablations in Section~\ref{sec:ablationcdp}, these qualitative observations are consistent with the intended role of degraded targets in reducing ambiguous cross-modal prediction while retaining useful structural supervision.
    
    \subsubsection{Representation Similarity under Varying Local Correspondence}

\begin{figure}[tbp]
        \centering
        \includegraphics[width=\linewidth]{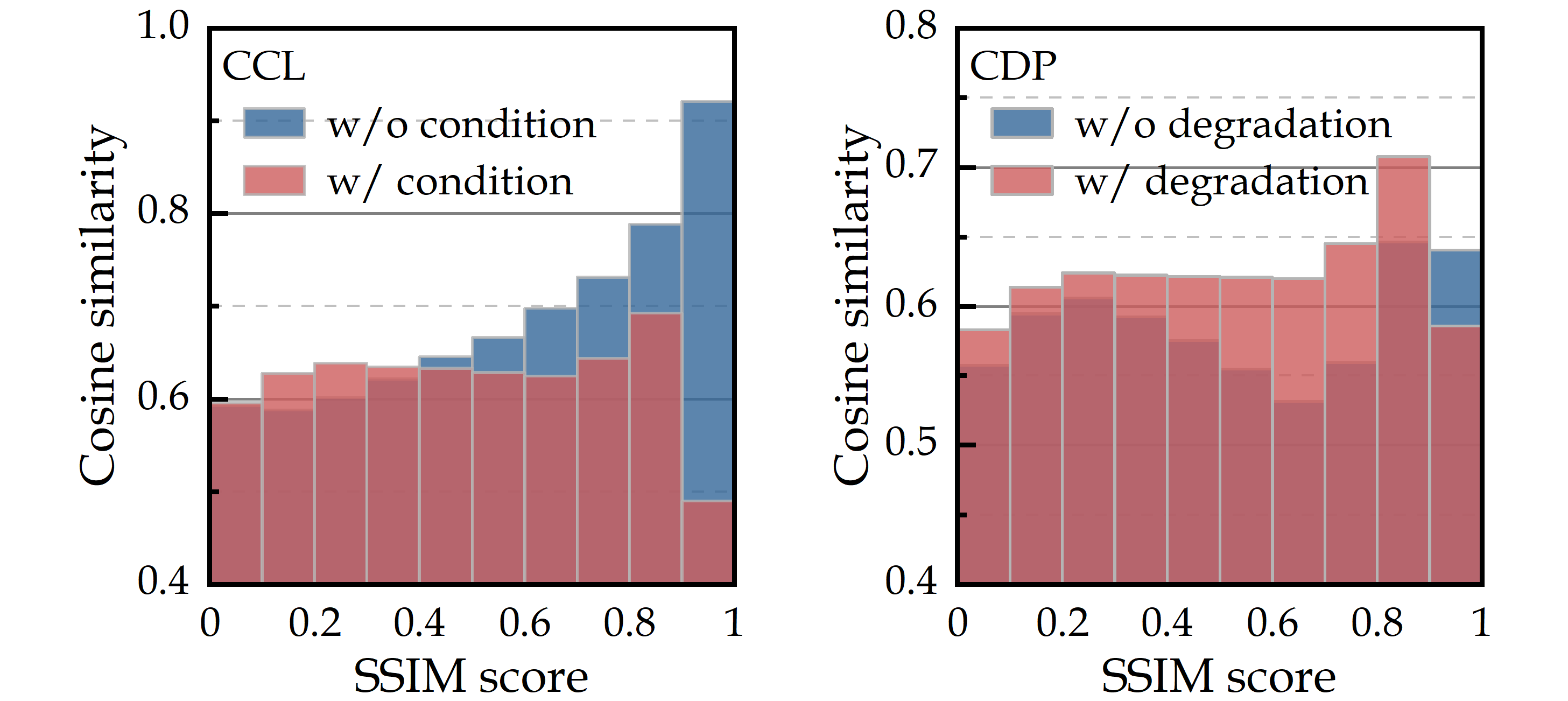}
        \vspace{-6mm}
        \caption{Raw-embedding similarity under varying local optical-SAR correspondence. Paired patches are grouped by their patch-level SSIM computed from the original optical-SAR inputs, which serves as an empirical descriptor of local visual correspondence. Cosine similarity between the corresponding optical and SAR encoder embeddings is averaged within each bin. Models trained with CCL exhibit a substantially weaker dependence of base-representation similarity on SSIM, while CDP produces a milder redistribution. All similarities are measured before cross-attention or prediction decoding.}
        \label{fig:analysis_align}
        \vspace{-2mm}
    \end{figure}

    We finally examine how the similarity of learned base representations varies with local optical-SAR correspondence. In Fig.~\ref{fig:analysis_align}, corresponding optical and SAR patches are grouped by their patch-level SSIM computed from the original inputs, which is used as an empirical descriptor of local visual correspondence. The vertical axis reports the cosine similarity between their raw encoder embeddings before the cross-attention module or prediction decoder.

    Relative to their respective direct counterparts, models trained with CCL and CDP exhibit a similar qualitative redistribution of raw-embedding similarity. In both cases, similarity decreases for the highest-SSIM pairs while generally increasing for lower-SSIM pairs. The effect is stronger for CCL: similarity decreases for most bins above SSIM 0.4, whereas CDP mainly reduces the peak in the highest-SSIM bin and increases similarity over the remaining range.

    This redistribution shows that the proposed objectives do not simply increase cross-modal similarity uniformly. Instead, they reduce the dependence of base-representation similarity on patch-level SSIM, narrowing the contrast between visually well-corresponded and more difficult optical-SAR pairs. Note that SSIM is used here only as an empirical descriptor of local visual correspondence, rather than as a direct measure of shared semantic information or the latent correspondence condition in Section~\ref{sec:theory}. The observed redistribution is also consistent with the UMAP visualization in Fig.~\ref{fig:analysis_umap}, where CCL and CDP show more coherently organized paired representations without collapsing the two modality distributions.

    \section{Conclusion}

    We study high-resolution optical-SAR joint pretraining under partial and pair-dependent cross-modal correspondence. Under simplified models, we show that a fixed alignment relation can be insufficient when correspondence varies across pairs and characterize ambiguity in cross-modal prediction. These observations motivate CoDe-MAE, which combines OKD with pair-conditioned alignment through CCL and degraded-target prediction through CDP.

    We curate OSPretrain-1M and use CoDe-MAE to pretrain MMATR-X, which achieves leading transfer performance across optical-only, SAR-only, and dual-modal tasks. Controlled optical-SAR experiments validate the method under matched settings, while optical-DEM and RGB-depth experiments provide additional evidence across heterogeneous modality pairs. The results highlight the importance of adapting alignment and prediction supervision to partial correspondence while retaining useful modality-specific variation.

    \bibliographystyle{IEEEtran}
    \normalem
    \bibliography{codeos_refs}

    \begin{IEEEbiography}[{\includegraphics[width=1in,height=1.25in,clip,keepaspectratio]{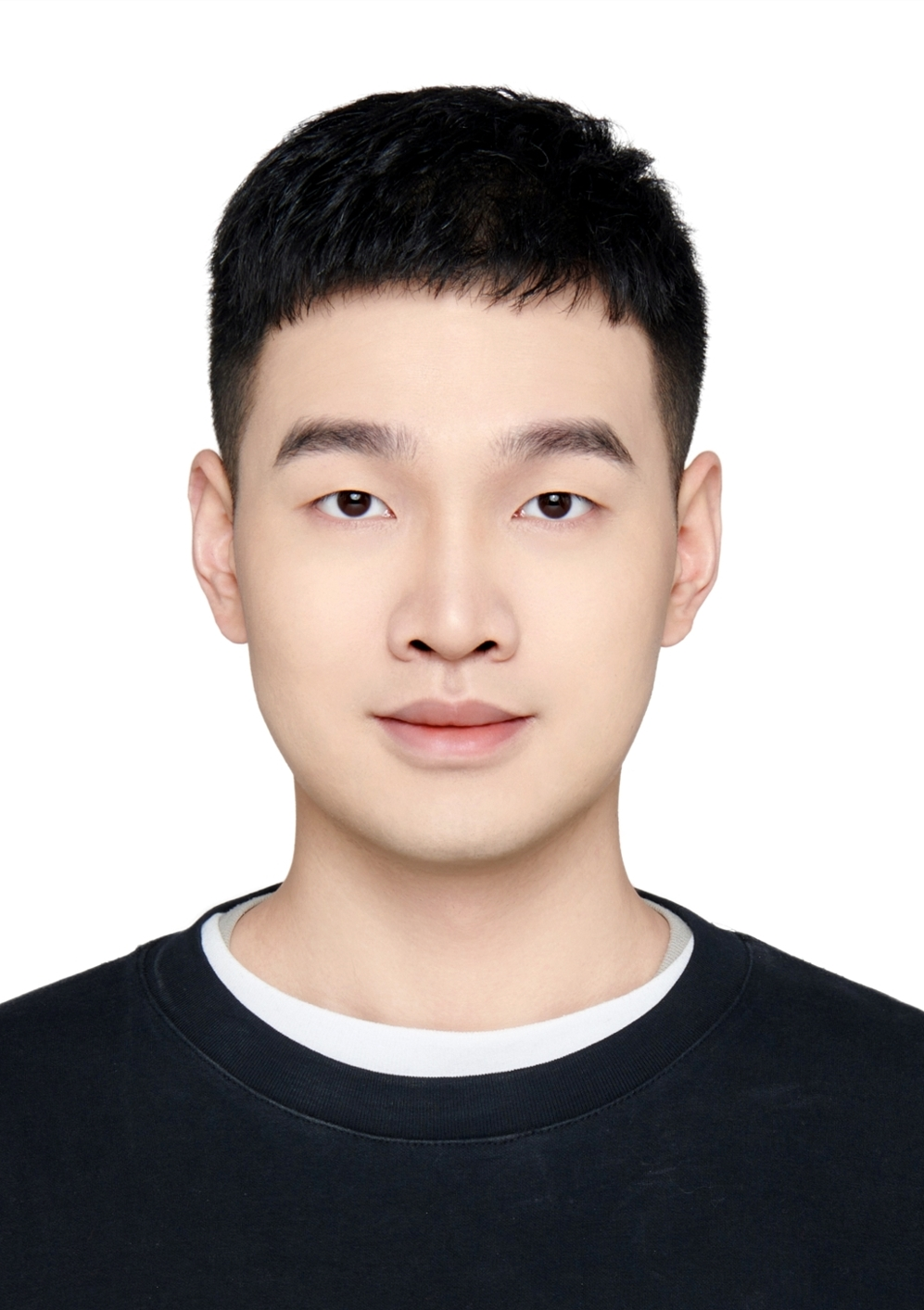}}]{Bowen Peng}
		received his B.S. and M.S. degrees from the National University of Defense Technology (NUDT), Changsha, China, respectively in 2020 and 2022. He is pursuing the Ph.D. degree with the College of Electronic Science and Technology, NUDT. His research interests include adversarial robustness of deep learning and trustworthy remote sensing object recognition.
	\end{IEEEbiography}
    
    \begin{IEEEbiography}[{\includegraphics[width=1in,height=1.25in,clip,keepaspectratio]{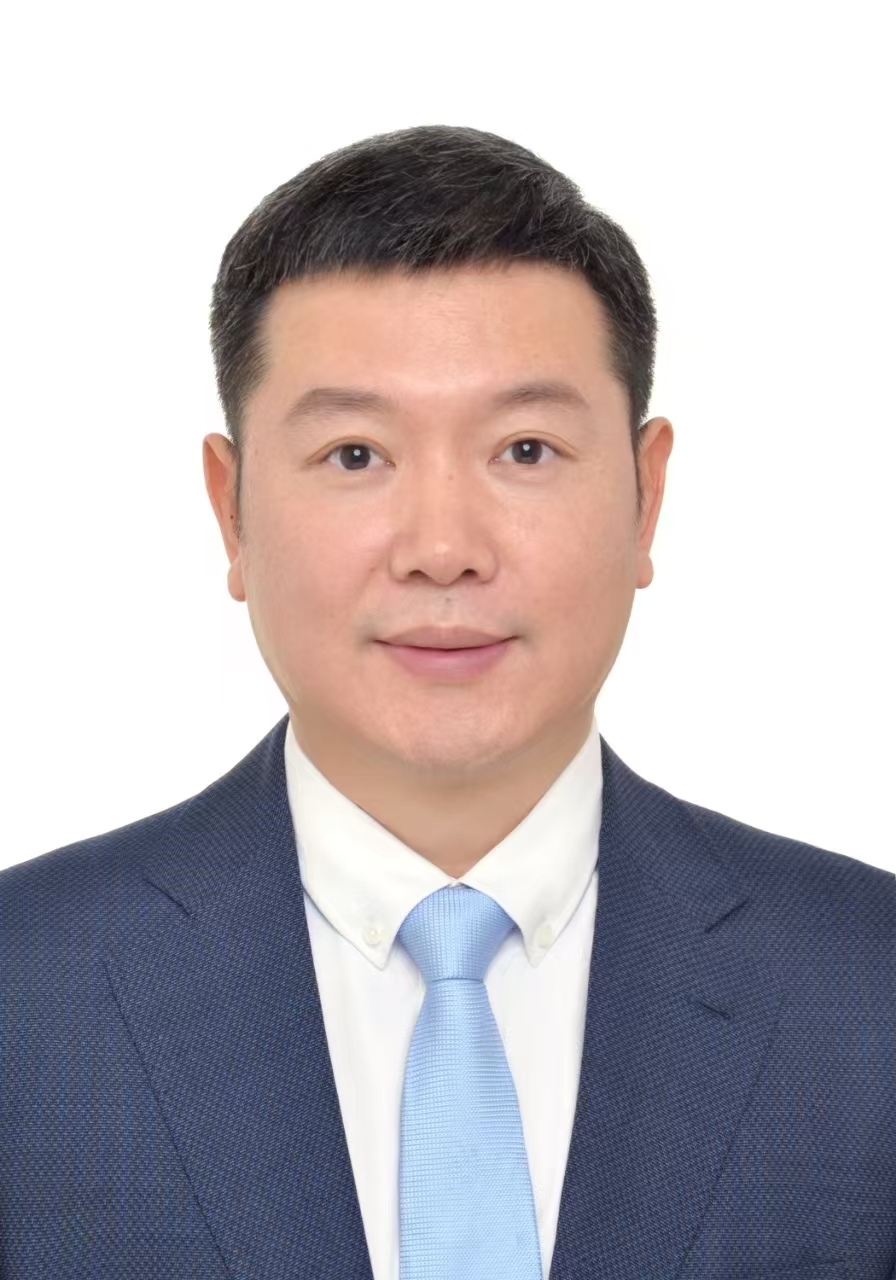}}]{Yongxiang Liu} received his Ph.D. degree in Information and Communication Engineering from the National University of Defense Technology (NUDT), Changsha, China, in 2004. Currently, He is a Full Professor in the College of Electronic Science and Technology, NUDT. His research interests mainly include remote sensing imagery analysis, radar signal processing, object recognition and Inverse SAR (ISAR) imaging, and machine learning. 
	\end{IEEEbiography}

    \begin{IEEEbiography}[{\includegraphics[width=1in,height=1.25in,clip,keepaspectratio]{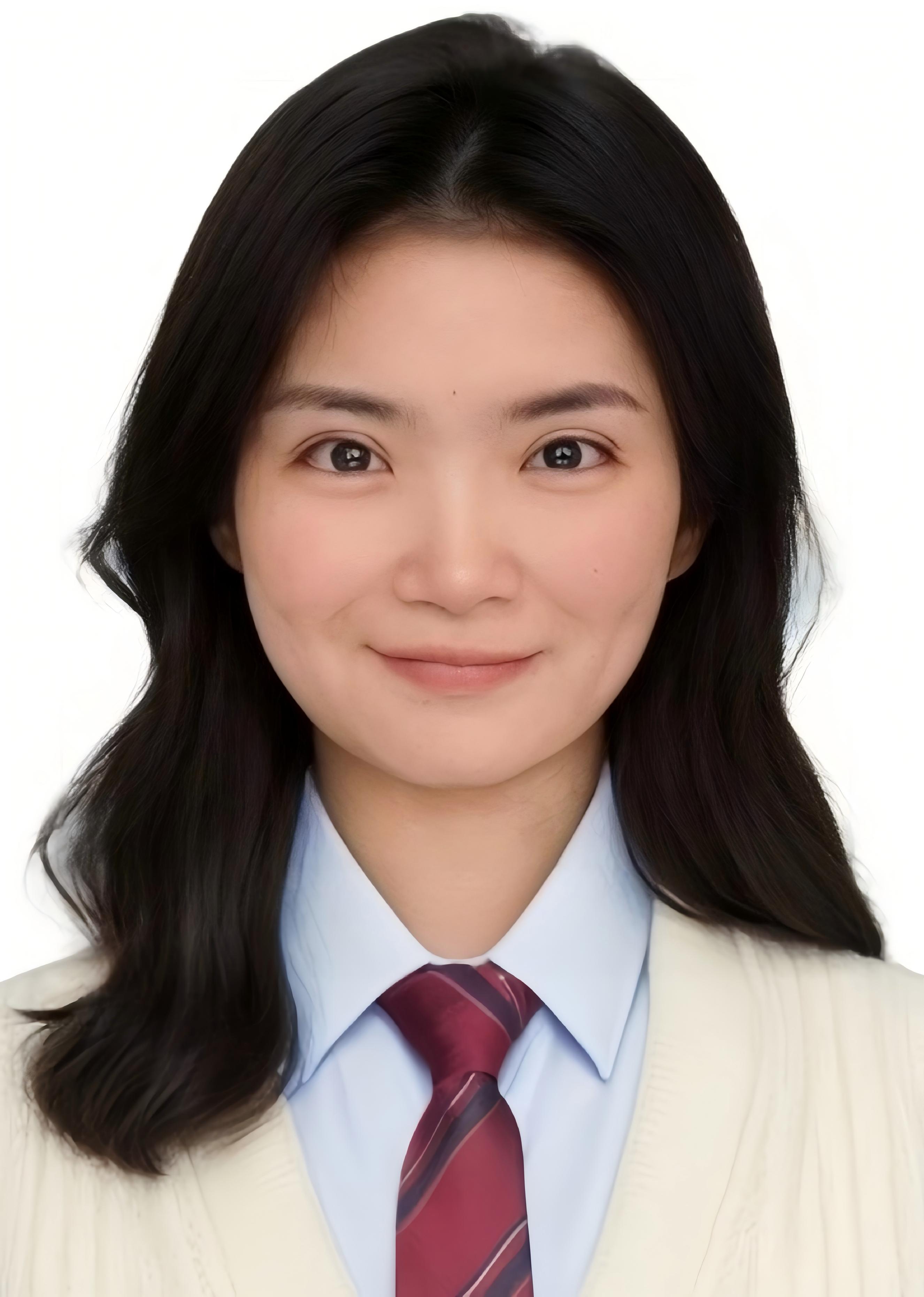}}]{Jie Zhou} received her B.S. degree from the National University of Defense Technology (NUDT), Changsha, China, in 2020, where she is currently pursuing the Ph.D. degree with the College of Electronic Science and Technology, NUDT. Her research interests include generative models, remote sensing image interpretation, and SAR target recognition.
	\end{IEEEbiography}

    \begin{IEEEbiography}[{\includegraphics[width=1in,height=1.25in,clip,keepaspectratio]{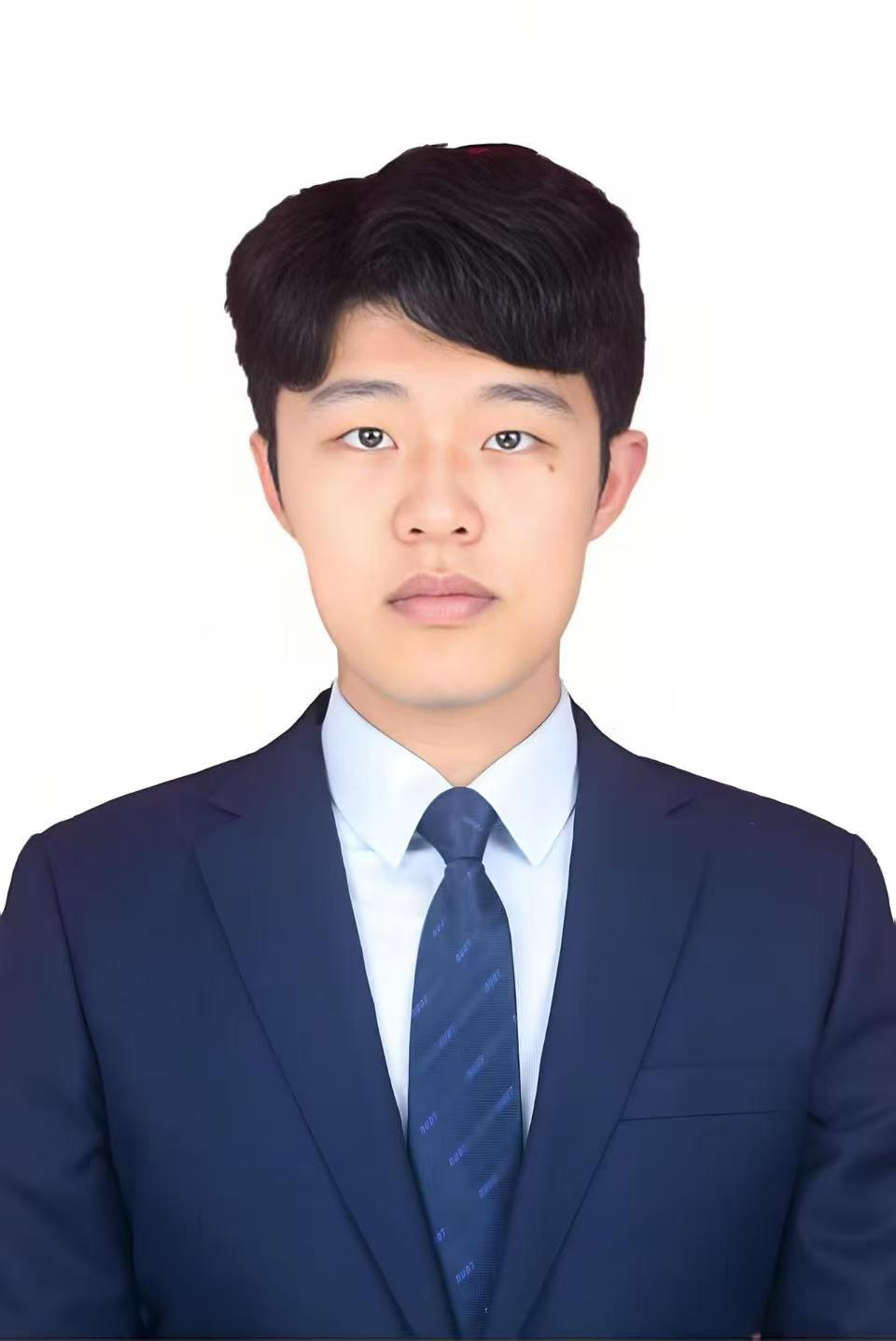}}]{Xiaodong Chen} received the university B.E. degree in physics form National University of Defense Technology, Changsha, China, in 2024. He is pursuing his Ph.D. degree at National University of Defense Technology, Changsha, China. His research interests mainly focused on multimodal target recognition and deep learning.
	\end{IEEEbiography}

    \begin{IEEEbiography}[{\includegraphics[width=1in,height=1.25in,clip,keepaspectratio]{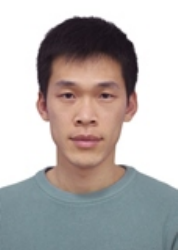}}]{Tienpeng Liu} received the B.Eng., M.Eng., and Ph.D. degrees from the National University of Defense Technology (NUDT), Changsha, China, in 2008, 2011, and 2016, respectively. He is currently a Professor with the College of Electronic Science and Technology, NUDT. His primary research interests include radar signal processing and automatic target recognition.
	\end{IEEEbiography}

    \begin{IEEEbiography}[{\includegraphics[width=1in,height=1.25in,clip,keepaspectratio]{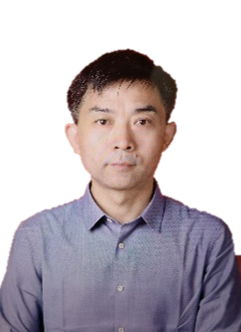}}]{Xiaogang Yu} received his Ph.D. degree in Communication and Information System from Beijing Institute of Technology (BIT), Beijing, China, in 2005. Currently, he is a Full Professor in Beijing Institute of Remote Sensing Information. His research interests mainly include remote sensing mission planning, multi-source remote sensing information processing, and deep learning.
	\end{IEEEbiography}
		
	\begin{IEEEbiography}[{\includegraphics[width=1in,height=1.25in,clip,keepaspectratio]{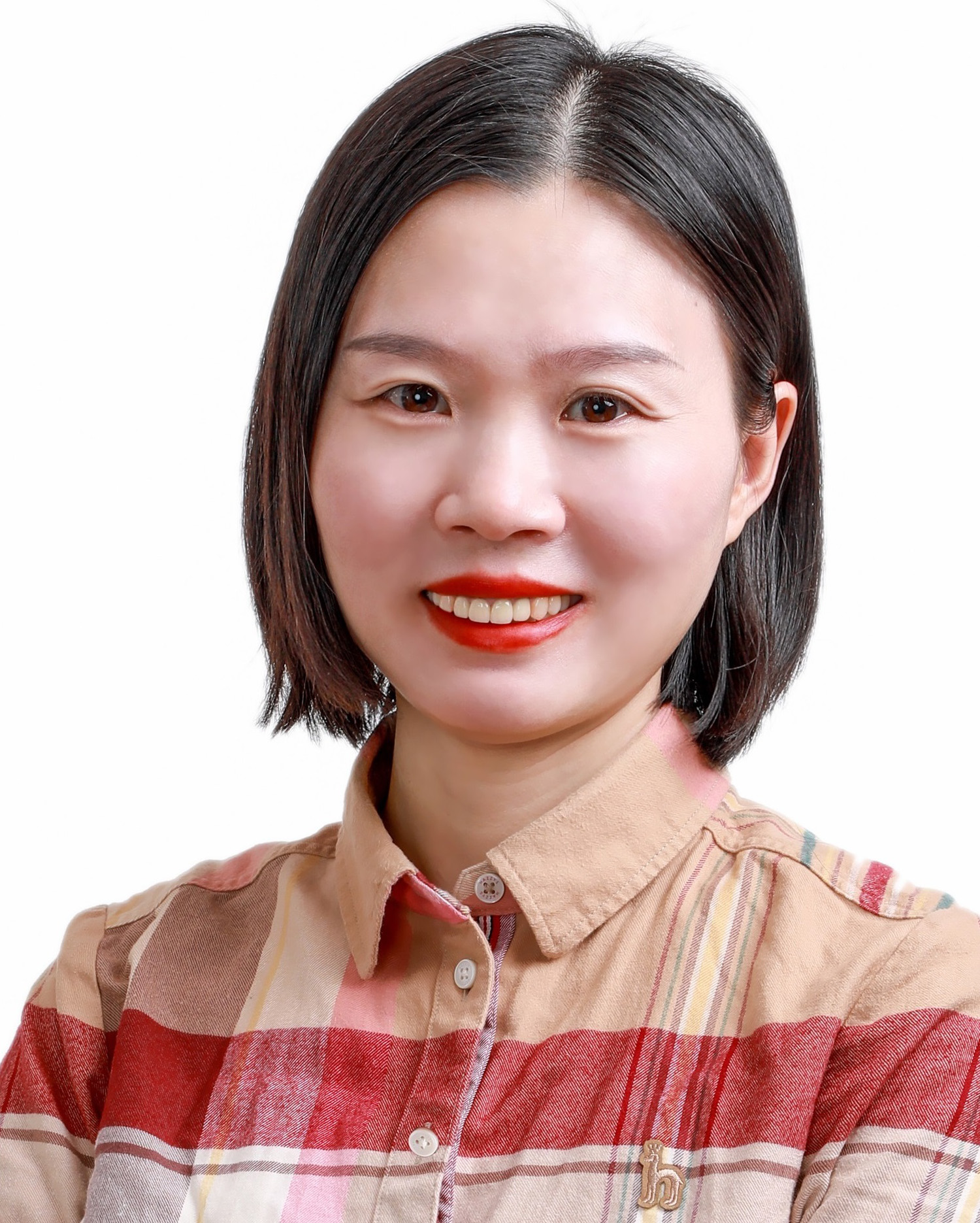}}]{Li Liu} (SM' 19) is a full professor with the National University of Defense Technology. She served as the leading guest editor of the special issues for IEEE TPAMI and IJCV and the Associate Editor for IEEE TCSVT. She currently serves as the Senior Associate Editor (SAE) for IEEE Transactions on Image Processing, the Associate Editor for IEEE Transactions on Multimedia and IEEE Transactions on Geoscience and Remote Sensing. Her current research interests include computer vision, pattern recognition,  machine learning and remote sensing image analysis. Her papers currently have over 27,000 citations in Google Scholar.
	\end{IEEEbiography}
	
    \vfill
		
	\end{document}